# Dynamic Active Constraints for Surgical Robots using Vector Field Inequalities

Murilo M. Marinho, *Member, IEEE,* Bruno V. Adorno, *Senior Member, IEEE,* Kanako Harada, *Member, IEEE*, and Mamoru Mitsuishi, *Member, IEEE*

*Abstract*—Robotic assistance allows surgeons to perform dexterous and tremor-free procedures, but robotic aid is still underrepresented in procedures with constrained workspaces, such as deep brain neurosurgery and endonasal surgery. In these procedures, surgeons have restricted vision to areas near the surgical tooltips, which increases the risk of unexpected collisions between the shafts of the instruments and their surroundings. In this work, our vector-field-inequalities method is extended to provide dynamic active-constraints to any number of robots and moving objects sharing the same workspace. The method is evaluated with experiments and simulations in which robot tools have to avoid collisions autonomously and in real-time, in a constrained endonasal surgical environment. Simulations show that with our method the combined trajectory error of two robotic systems is optimal. Experiments using a real robotic system show that the method can autonomously prevent collisions between the moving robots themselves and between the robots and the environment. Moreover, the framework is also successfully verified under teleoperation with tool–tissue interactions.

*Index Terms*—virtual fixtures; collision avoidance; dual quaternions; optimization-based control

## I. Introduction

**M**ICROSURGERY requires surgeons to operate with submillimeter precision while handling long thin tools and viewing the workspace through an endoscope or a microscope. For instance, in endonasal and deep neurosurgery, surgeons use a pair of instruments with a length of 100-300 mm and a diameter of 3-4 mm. Furthermore, in endonasal surgery images are obtained with a 4-mm-diameter endoscope, whereas deep neurosurgery requires a microscope. The workspace in both cases can be approximated by a truncated cone with a length of 80-110 mm and a diameter of 20-30 mm [1], [2].

This research was funded in part by the ImPACT Program of the Council for Science, Technology and Innovation (Cabinet Office, Government of Japan) and in part by the JSPS KAKENHI Grant Number JP19K14935.

Murilo M. Marinho, Kanako Harada, and Mamoru Mitsuishi are with the Department of Mechanical Engineering, the University of Tokyo, Tokyo, Japan. Email:`{murilo, kanako, mamoru}@nml.t.u-tokyo.ac.jp`. Murilo M. Marinho has been supported by the Japanese Ministry of Education, Culture, Sports, Science, and Technology (MEXT).

Bruno V. Adorno is with the Department of Electrical Engineering and with the Graduate Program in Electrical Engineering - Federal University of Minas Gerais, Belo Horizonte-MG, Brazil. Email: `adorno@ufmg.br`. Bruno V. Adorno has been supported by the Brazilian agencies CAPES, CNPq, FAPEMIG, and by the project INCT (National Institute of Science and Technology) under the grant CNPq (Brazilian National Research Council) 465755/2014-3, FAPESP (São Paulo Research Foundation) 2014/50851-0.

In hands-on microsurgery, the surgeon is fully aware of the tools' positions with respect to the workspace and is able to feel forces to prevent damage to structures. However, the constant collisions between surrounding tissues and other instruments reduce the surgeon's dexterity and increase the difficulty of the task. As vision is often limited to a region near the surgical tool tips, it is difficult for surgeons to compensate for the restrictions induced by unseen collision points. Moreover, disturbances caused by hand tremor may be larger than the structures being treated, which is problematic, as accuracy is paramount in these surgical procedures. In this context, robots are used as assistive equipment to increase accuracy and safety and to reduce hand tremors and mental load  [1]–[6].

To increase accuracy and attenuate hand tremors, most surgical robots are commanded in task space coordinates through either teleoperation [1], [3]–[8] or comanipulation [9], [10]. Motion commands go through a scaling and/or filtering stage, whose output is used to generate joint space control inputs by using a kinematic control law based on the robot's differential kinematics [11]. Kinematic control laws are valid when low accelerations are imposed in the joint space and are ubiquitous in control algorithms designed for surgical robotics [7], [8], [12], [13] since low accelerations are expected in those scenarios. In microsurgery, slow and low amplitude motions are the rule.

Together with such kinematic control laws, active constraints (virtual fixtures) can be used in order to provide an additional layer of safety to keep the surgical tool from entering a restricted region or from leaving a safe region, even if the robot is commanded otherwise [12], [13]. A thorough survey on active constraints was done by Bowyer *et al.* [14]. More recent papers published after that survey presented the use of guidance virtual fixtures to assist with knot tying in robotic laparoscopy [15] and virtual fixtures to allow surgeons to feel the projected force feedback from the distal end of a tool to the proximal end of a tool in a comanipulation context [16].

In real applications, some of the objects in the surgical workspace are dynamic. Usually these elements are other tools sharing the same restricted workspace or moving organs. Most of the literature on dynamic active constraints regards the latter and, more specifically, the development of techniques to avoid collisions between the tool tip and the beating heart [17]–[20]. For instance, Gibo *et al.* [17] studied the proxy method with moving forbidden zones, Ryden and Chizeck [20] proposed virtual fixtures for dynamic restricted zones using point clouds, and Ren *et al.* [21] proposed dynamic active constraints using medical images to build potential fields, all aiming to reduce the contact force between the tool tip and the beating-heart's surface. A work that considered the entire robot instead of just the tool tip was proposed by Kwok *et al.* [22], who developed a control framework for a snake robot based on an optimization problem. They used dynamic active constraints in the context of beating heart surgery [18] and guidance virtual fixtures [23], both of which were specialized for snake robots.

### A. Enforcement of constraints

The literature on static and moving virtual fixtures is generally focused on the interactions between the tool tip and the environment. In these cases, generating force feedback on the master side is quite straightforward and may be sufficient to avoid damaging healthy tissue. In microsurgery, however, interactions also occur far from the tool tips. Indeed, our long-standing cooperation with medical doctors [1], [2], [4]–[6] indicates that, as the workspace is highly constrained, surgical tools may suffer collisions along their structures, whose effects on the tool tip can be complex; therefore, effectively projecting those collisions to the master interface for effective force feedback is challenging.

Other forms of feedback, such as visual cues [6], [24], that warn the user about the proximity to restricted regions can be ineffective when using redundant robotic systems [1] since the mapping between the master interface and the robot configuration is not one to one. In such cases, moving the tool tip in a given direction may correspond to an infinite number of configurations due to redundancy, which can be counterintuitive for the surgeon as the same tool tip trajectory can correspond to collision-free motions or motions with collision depending on the robot configuration. In fact, it is safer and simpler if the robot avoids restricted zones—especially those outside the field of view—autonomously. In this way, the increasing prevalence of redundant robots is turned into an advantage if used together with a proper control framework as the one proposed in this work.

### B. Surgical robot design and geometry

There have been many proposed designs for surgical robotics for constrained workspaces, such as rigid-link robots [1]–[3], [5], snake-like robots [18], [22], [23], and flexible robots [7], [8], [25], [26]. In this regard, a suitable framework for the generation of dynamic active constraints should not be limited by a given robot geometry, and should be as generalizable as possible as the one proposed in this work.

### C. Related work

To the best of the authors' knowledge, the framework in the medical robotics literature that more closely meets microsurgical requirements is the one initially developed by Funda *et al.* [27], which uses quadratic programming. Their framework does not require a specific robot geometry; it can handle points in the robot besides the tool tip and also considers equality and inequality constraints, which gives hard distance limits between any point in the manipulator and objects in the environment. Their framework was extended by Kapoor *et al.* [12], who developed a library of virtual fixtures using five task primitives that can be combined into customized active constraints using nonlinear constraints or linear *approximations*. Li *et al.* [13] used the extended framework to aid operators in moving, without collision, a single surgical tool in a highly constrained space in sinus surgery; however, owing to how the obstacles were modeled using a covariance tree, which requires a long computational time, dynamic virtual fixtures cannot be considered [21], [22].

In prior approaches [12], [13], [17]–[20], [22], [23], [27], obstacle constraints are activated whenever the robot reaches the boundary of a restricted zone, which might result in acceleration peaks. Some authors attempted to address this issue. For instance, Xia *et al.* [28] reduced the proportional gain in an admittance control law in proportion to the distance between the tool tip and the nearest obstacle. This allowed the system to smoothly avoid collisions but also impeded motion tangential to the obstacle boundary. Prada and Payandeh [29] proposed adding a time-varying gain to smooth the attraction force of attractive virtual fixtures.

Outside the literature on active constraints/virtual fixtures, Faverjon and Tournassoud [30] proposed a path planner based on velocity dampers, which are an alternative to potential fields, to allow the generation of restricted zones without affecting tangential motion. Their technique was evaluated in simulations of a manipulator navigating through a nuclear reactor. To reduce computational costs, their method used the *approximate* distance between convex solids obtained through iterative optimization. Kanoun *et al.* [31] extended velocity dampers to a task-priority framework. Their technique was validated with a simulated humanoid robot.

Finding the exact distance and the time-derivative of the distance between different relevant primitives can increase the computational speed and accuracy. This is of high importance in microsurgical applications, as properly using the constrained workspace might require the tool shafts to get arbitrarily close to each other without collision, in addition to the fact that all calculations must be performed in real time.

### D. Statement of contributions

In this work, we further extend the research done in [32]. First, we extend the framework to take into consideration dynamic active constraints, which may include any number of robotic systems sharing the same workspace as well as dynamic objects whose velocity can be tracked or estimated. The distance and its derivative between elements is found algebraically, allowing accurate and fast computation of the constraints, which is paramount for real-time evaluation in microsurgical settings. In addition, we introduce more relevant primitives that are coupled to the robots in order to enrich the description of the constraints. Second, we evaluate the dynamic-active-constraint framework in in-depth simulation studies, which demonstrate the advantages of dynamic active constraints over static ones. Third, the framework is implemented in a physical system composed of a pair of robotic manipulators. Experiments are performed, the first of which is an autonomous tool tip tracing experiment using a realistic endonasal surgical setup, with an anatomically correct 3D printed head model, in which active constraints are used to avoid collisions. Lastly, using a similar set of constraints, a teleoperation experiment is performed in which the robot is used to cut a flexible membrane.

In contrast with prior techniques in the medical robotics literature [12], [13], [17]–[20], [22], [23], [27], our technique provides smooth velocities at the boundaries of restricted zones. Moreover, it does not disturb or prevent tangential velocities, which is different from [28] and [29]. When specifically considering the literature on dynamic active constraints, our technique considers an arbitrary number of simultaneous frames in the robot and not only the tool tip, as in [17]–[20]; furthermore, it is applicable to general robot geometries, in contrast to solutions for snake robots [18], [22], [23]. Outside medical robotics applications, our framework is a generalization of velocity dampers [30].

*E. Organization of the paper*

Section II introduces the required mathematical background; more specifically, it presents a review of dual-quaternion algebra, differential kinematics, and quadratic programming. Section III explains the proposed vector-field inequalities and Section IV introduces primitives used to model the relation between points, lines, planes, and to enforce the task's geometrical constraints. To bridge the gap between theory and implementation, a computational algorithm and an example of how to combine the proposed primitives into relevant safe zones are shown in Section V. Section VI presents simulations to evaluate different combinations of dynamic and static active constraints, in addition to experiments in a real platform to evaluate the exponential behavior of the velocity constraint towards a static plane, and two experiments in a realistic environment for endonasal surgery. Finally, Section VII concludes the paper and presents suggestions for future work.

## II. MATHEMATICAL BACKGROUND

The proposed virtual-fixtures framework makes extensive use of dual-quaternion algebra thanks to several advantages over other representations. For instance, unit dual quaternions do not have representational singularities and are more compact and computationally efficient than homogeneous transformation matrices [33]. Their strong algebraic properties allow different robots to be systematically modeled [33]–[36]. Furthermore, dual quaternions can be used to represent rigid motions, twists, wrenches, and several geometrical primitives, e.g., Plücker lines, planes, and cylinders, in a very straightforward way [37]. The next subsection introduces the basic definitions of quaternions and dual quaternions; more information can be found in [34], [37], [38].

*A. Quaternions and dual quaternions*

Quaternions can be regarded as an extension of complex numbers. The quaternion set is

$$\mathbb{H} \triangleq \left\{ h_1 + \hat{\imath} h_2 + \hat{\jmath} h_3 + \hat{k} h_4 \, : \, h_1, h_2, h_3, h_4 \in \mathbb{R} \right\},$$

TABLE I
MATHEMATICAL NOTATION.

| Notation | Meaning |
|---|---|
| $\mathbb{H}, \mathbb{H}_p$ | Set of quaternions and pure quaternions (the latter is isomorphic to $\mathbb{R}^3$ under addition) |
| $\mathcal{H}, \mathcal{H}_p$ | Set of dual quaternions and pure dual quaternions (the latter is isomorphic to $\mathbb{R}^6$ under addition) |
| $\mathbb{S}^3, \underline{\mathcal{S}}$ | Set of unit quaternions and unit dual quaternions |
| $\boldsymbol{p} \in \mathbb{H}_p$ | position of a point in space with known first order kinematics |
| $\underline{\boldsymbol{l}} \in \mathcal{H}_p \cap \underline{\mathcal{S}}$ | line in space with known first order kinematics |
| $\underline{\boldsymbol{\pi}}$ | plane in space with known first order kinematics |
| $\boldsymbol{q} \in \mathbb{R}^n$ | vector of joints' configurations |
| $\boldsymbol{J}_\# \in \mathbb{R}^{m \times n}$ | Jacobian relating joint velocities and an $m$−degrees-of-freedom task $\#$ |
| $\boldsymbol{t} \in \mathbb{H}_p$ | position of a point in the robot |
| $\boldsymbol{r} \in \mathbb{S}^3$ | orientation of a frame in the robot |
| $\underline{\boldsymbol{x}} \in \underline{\mathcal{S}}$ | pose of a frame in the robot |
| $\underline{\boldsymbol{l}}_z \in \mathcal{H}_p \cap \underline{\mathcal{S}}$ | line passing through the $z$-axis of a frame in the robot |
| $d_{\#1, \#2}$ | distance function between geometric entities $\#1$ and $\#2$ |
| $D_{\#1, \#2}$ | squared distance between geometric entities $\#1$ and $\#2$ |
| $\eta$ | task-space proportional gain |
| $\eta_d, \eta_D$ | gains for the distance or squared distance constraints |

in which the imaginary units $\hat{\imath}$, $\hat{\jmath}$, and $\hat{k}$ have the following properties: $\hat{\imath}^2 = \hat{\jmath}^2 = \hat{k}^2 = \hat{\imath}\hat{\jmath}\hat{k} = -1$. The dual quaternion set is

$$\mathcal{H} \triangleq \left\{ \boldsymbol{h} + \varepsilon \boldsymbol{h}' \, : \, \boldsymbol{h}, \boldsymbol{h}' \in \mathbb{H}, \, \varepsilon^2 = 0, \, \varepsilon \neq 0 \right\},$$

where $\varepsilon$ is the dual (or Clifford) unit [38]. Addition and multiplication are defined for dual quaternions in an analogous manner as those for complex numbers; hence, we just need to respect the properties of the imaginary and dual units.

Given $\underline{\boldsymbol{h}} \in \mathcal{H}$ such that $\underline{\boldsymbol{h}} = h_1 + \hat{\imath} h_2 + \hat{\jmath} h_3 + \hat{k} h_4 + \varepsilon \left( h_1' + \hat{\imath} h_2' + \hat{\jmath} h_3' + \hat{k} h_4' \right)$, we define the operators

$$\mathcal{P}\left(\underline{\boldsymbol{h}}\right) \triangleq h_1 + \hat{\imath} h_2 + \hat{\jmath} h_3 + \hat{k} h_4, \quad \mathcal{D}\left(\underline{\boldsymbol{h}}\right) \triangleq h_1' + \hat{\imath} h_2' + \hat{\jmath} h_3' + \hat{k} h_4',$$

and

$$\operatorname{Re}\left(\underline{\boldsymbol{h}}\right) \triangleq h_1 + \varepsilon h_1',$$
$$\operatorname{Im}\left(\underline{\boldsymbol{h}}\right) \triangleq \hat{\imath} h_2 + \hat{\jmath} h_3 + \hat{k} h_4 + \varepsilon \left( \hat{\imath} h_2' + \hat{\jmath} h_3' + \hat{k} h_4' \right).$$

The conjugate of $\underline{\boldsymbol{h}}$ is $\underline{\boldsymbol{h}}^* \triangleq \operatorname{Re}\left(\underline{\boldsymbol{h}}\right) - \operatorname{Im}\left(\underline{\boldsymbol{h}}\right)$, and its norm is given by $\|\underline{\boldsymbol{h}}\| = \sqrt{\underline{\boldsymbol{h}} \underline{\boldsymbol{h}}^*} = \sqrt{\underline{\boldsymbol{h}}^* \underline{\boldsymbol{h}}}$.

The set $\mathbb{H}_p \triangleq \{\boldsymbol{h} \in \mathbb{H} \, : \, \operatorname{Re}\left(\boldsymbol{h}\right) = 0\}$ has a bijective relation with $\mathbb{R}^3$. This way, the quaternion $\left(x\hat{\imath} + y\hat{\jmath} + z\hat{k}\right) \in \mathbb{H}_p$ represents the point $(x, y, z) \in \mathbb{R}^3$. The set of quaternions with a unit norm is $\mathbb{S}^3 \triangleq \{\boldsymbol{h} \in \mathbb{H} \, : \, \|\boldsymbol{h}\| = 1\}$, and $\boldsymbol{r} \in \mathbb{S}^3$ can always be written as $\boldsymbol{r} = \cos\left(\phi/2\right) + \boldsymbol{v} \sin\left(\phi/2\right)$, where $\phi \in \mathbb{R}$ is the rotation angle around the rotation axis $\boldsymbol{v} \in \mathbb{S}^3 \cap \mathbb{H}_p$ [37]. The elements of the set $\underline{\mathcal{S}} \triangleq \{\underline{\boldsymbol{h}} \in \mathcal{H} \, : \, \|\underline{\boldsymbol{h}}\| = 1\}$ are called unit dual quaternions and represent the tridimensional poses (i.e., the combined position and orientation) of rigid bodies. Given $\underline{\boldsymbol{x}} \in \underline{\mathcal{S}}$, it can always be written as $\underline{\boldsymbol{x}} = \boldsymbol{r} + \varepsilon \left(1/2\right) \boldsymbol{t} \boldsymbol{r}$, where $\boldsymbol{r} \in \mathbb{S}^3$ and $\boldsymbol{t} \in \mathbb{H}_p$ represent the orientation and position, respectively [38]. The set $\underline{\mathcal{S}}$ equipped with the multiplication operation forms the group $\operatorname{Spin}(3) \ltimes \mathbb{R}^3$, which double covers $\operatorname{SE}(3)$.

Elements of the set $\mathcal{H}_p \triangleq \{\underline{\boldsymbol{h}} \in \mathcal{H} \, : \, \operatorname{Re}\left(\underline{\boldsymbol{h}}\right) = 0\}$ are called pure dual quaternions. Given $\underline{\boldsymbol{a}}, \underline{\boldsymbol{b}} \in \mathcal{H}_p$, the inner product and cross product are respectively [34], [37]

$$\langle \underline{\boldsymbol{a}}, \underline{\boldsymbol{b}} \rangle \triangleq -\frac{\underline{\boldsymbol{a}}\underline{\boldsymbol{b}} + \underline{\boldsymbol{b}}\underline{\boldsymbol{a}}}{2}, \qquad \underline{\boldsymbol{a}} \times \underline{\boldsymbol{b}} \triangleq \frac{\underml{\boldsymbol{a}}\underline{\boldsymbol{b}} - \underline{\boldsymbol{b}}\underline{\boldsymbol{a}}}{2}. \qquad (1)$$

The operator $\text{vec}_4$ maps quaternions into $\mathbb{R}^4$, and $\text{vec}_8$ maps dual quaternions into $\mathbb{R}^8$. For instance, $\text{vec}_4\, \boldsymbol{h} \triangleq \begin{bmatrix} h_1 & h_2 & h_3 & h_4 \end{bmatrix}^T$, and $\text{vec}_8\, \underline{\boldsymbol{h}} \triangleq \begin{bmatrix} h_1 & h_2 & h_3 & h_4 & h'_1 & h'_2 & h'_3 & h'_4 \end{bmatrix}^T$.

Given $\boldsymbol{h}_1, \boldsymbol{h}_2 \in \mathbb{H}$, the Hamilton operators are matrices that satisfy $\text{vec}_4\,(\boldsymbol{h}_1 \boldsymbol{h}_2) = \overset{+}{\boldsymbol{H}}_4(\boldsymbol{h}_1) \text{vec}_4\, \boldsymbol{h}_2 = \overline{\boldsymbol{H}}_4(\boldsymbol{h}_2)\text{vec}_4\, \boldsymbol{h}_1$. Analogously, given $\underline{\boldsymbol{h}}_1, \underline{\boldsymbol{h}}_2 \in \mathcal{H}$, the Hamilton operators satisfy $\text{vec}_8\,(\underline{\boldsymbol{h}}_1 \underline{\boldsymbol{h}}_2) = \overset{+}{\boldsymbol{H}}_8(\underline{\boldsymbol{h}}_1) \text{vec}_8\, \underline{\boldsymbol{h}}_2 = \overline{\boldsymbol{H}}_8(\underline{\boldsymbol{h}}_2)\text{vec}_8\, \underline{\boldsymbol{h}}_1$ [37].

From (1), one can find by direct calculation that the inner product of any two pure quaternions $\boldsymbol{a} = \hat{\imath} a_2 + \hat{\jmath} a_3 + \hat{k} a_4$ and $\boldsymbol{b} = \hat{\imath} b_2 + \hat{\jmath} b_3 + \hat{k} b_4$ is a real number given by

$$\langle \boldsymbol{a}, \boldsymbol{b} \rangle = -\frac{\boldsymbol{a}\boldsymbol{b} + \boldsymbol{b}\boldsymbol{a}}{2} = \text{vec}_4\, \boldsymbol{a}^T \text{vec}_4\, \boldsymbol{b} = \text{vec}_4\, \boldsymbol{b}^T \text{vec}_4\, \boldsymbol{a}. \quad (2)$$

Furthermore, the cross product between $\boldsymbol{a}$ and $\boldsymbol{b}$ is mapped into $\mathbb{R}^4$ as

$$\text{vec}_4\,(\boldsymbol{a} \times \boldsymbol{b}) = \underbrace{\begin{bmatrix} 0 & 0 & 0 & 0 \\ 0 & 0 & -a_4 & a_3 \\ 0 & a_4 & 0 & -a_2 \\ 0 & -a_3 & a_2 & 0 \end{bmatrix}}_{\overline{\boldsymbol{S}}(\boldsymbol{a})} \text{vec}_4\, \boldsymbol{b}$$
$$= \overline{\boldsymbol{S}}(\boldsymbol{a}) \text{vec}_4\, \boldsymbol{b} = \overline{\boldsymbol{S}}(\boldsymbol{b})^T \text{vec}_4\, \boldsymbol{a}. \quad (3)$$

When it exists, the time derivative of the squared norm of a time-varying quaternion $\boldsymbol{h}(t) \in \mathbb{H}_p$ is given by

$$\frac{d}{dt}\left(\|\boldsymbol{h}\|^2\right) = \dot{\boldsymbol{h}} \boldsymbol{h}^* + \boldsymbol{h} \dot{\boldsymbol{h}}^* = 2\langle \dot{\boldsymbol{h}}, \boldsymbol{h} \rangle. \quad (4)$$

*Using other rigid-body motion representations:* The technique proposed in this paper is based on dual quaternion algebra, but can also be used in conjunction with existing robotic systems using other rigid-body motion representations (e.g. homogenous transformation matrices), if required. The preferred approach is to use conversion functions to transform the input/output of those systems from/to the other representation to/from unit dual quaternions. For instance, the conversion can be based on the transformation between a rotation matrix and a quaternion [11].[1]

### B. Differential kinematics

Differential kinematics is the relation between task-space and joint-space velocities in the general form

$$\dot{\boldsymbol{x}} = \boldsymbol{J} \dot{\boldsymbol{q}},$$

in which $\boldsymbol{q} \triangleq \boldsymbol{q}(t) \in \mathbb{R}^n$ is the vector of the manipulator joints' configurations, $\boldsymbol{x} \triangleq \boldsymbol{x}(\boldsymbol{q}) \in \mathbb{R}^m$ is the vector of $m$ task-space variables, and $\boldsymbol{J} \triangleq \boldsymbol{J}(\boldsymbol{q}) \in \mathbb{R}^{m \times n}$ is a Jacobian matrix.

The task-space variables are the variables relevant for the specific task to be performed in any frame kinematically coupled to the robot. In many relevant robotic tasks, this means pose, position, or orientation control of the robot's end effector. For instance, if the pose of an arbitrary frame attached to the robot is written as $\underline{\boldsymbol{x}} \triangleq \underline{\boldsymbol{x}}(\boldsymbol{q}) \in \text{Spin}(3) \ltimes \mathbb{R}^3$, the corresponding differential kinematics is given by

$$\text{vec}_8\, \underline{\dot{\boldsymbol{x}}} = \boldsymbol{J}_{\underline{\boldsymbol{x}}} \dot{\boldsymbol{q}}, \quad (5)$$

where $\boldsymbol{J}_{\underline{\boldsymbol{x}}} \in \mathbb{R}^{8 \times n}$ is the dual quaternion analytical Jacobian, which can be found using dual quaternion algebra [33]. Similarly, given the position of a frame in the robot $\boldsymbol{t} \triangleq \boldsymbol{t}(\boldsymbol{q}) \in \mathbb{H}_p$ and the orientation of a frame in the robot $\boldsymbol{r} \triangleq \boldsymbol{r}(\boldsymbol{q}) \in \mathbb{S}^3$ such that $\underline{\boldsymbol{x}} = \boldsymbol{r} + \varepsilon (1/2) \boldsymbol{t}\boldsymbol{r}$, we have

$$\text{vec}_4\, \dot{\boldsymbol{t}} = \boldsymbol{J}_t \dot{\boldsymbol{q}}, \quad (6)$$
$$\text{vec}_4\, \dot{\boldsymbol{r}} = \boldsymbol{J}_r \dot{\boldsymbol{q}}, \quad (7)$$

where $\boldsymbol{J}_t, \boldsymbol{J}_r \in \mathbb{R}^{4 \times n}$ are also calculated from $\boldsymbol{J}_{\underline{\boldsymbol{x}}}$ using dual quaternion algebra [39].

In this work we extend the concept of controlling the pose, position, and orientation of the frames in the robot to also controlling the distances to points, lines, and planes. Henceforth, points, lines, and planes kinematically coupled to the robot are called *robot entities*.

### C. Quadratic programming for differential inverse kinematics[2]

In closed-loop differential inverse kinematics, we first define a task-space target $\boldsymbol{x}_d$ and task error $\tilde{\boldsymbol{x}} = \boldsymbol{x} - \boldsymbol{x}_d$. Considering $\dot{\boldsymbol{x}}_d = \boldsymbol{0}\ \forall t$ and a gain $\eta \in (0, \infty)$, the analytical solution to the convex optimization problem [40]

$$\min_{\dot{\boldsymbol{q}}} \|\boldsymbol{J}\dot{\boldsymbol{q}} + \eta \tilde{\boldsymbol{x}}\|_2^2, \quad (8)$$

is the set of minimizers[3]

$$Q \triangleq \left\{ \dot{\boldsymbol{q}}^o \in \mathbb{R}^n : \dot{\boldsymbol{q}}^o = -\boldsymbol{J}^\dagger \eta \tilde{\boldsymbol{x}} + \left(\boldsymbol{I} - \boldsymbol{J}^\dagger \boldsymbol{J}\right)\boldsymbol{z} \right\},$$

in which $\boldsymbol{J}^\dagger$ is the Moore–Penrose inverse of $\boldsymbol{J}$, $\boldsymbol{I}$ is an identity matrix of proper size, and $\boldsymbol{z} \in \mathbb{R}^n$ is an arbitrary vector. In particular, the analytical solution $\dot{\boldsymbol{q}}^o = -\boldsymbol{J}^\dagger \eta \tilde{\boldsymbol{x}}$ is the solution to the following optimization problem:

$$\min_{\dot{\boldsymbol{q}} \in Q} \|\dot{\boldsymbol{q}}\|_2^2. \quad (9)$$

Adding linear constraints to Problem 8 turns it into a linearly constrained quadratic programming problem requiring a numerical solver [41]. The standard form of Problem 8 with *linear* constraints is

$$\min_{\dot{\boldsymbol{q}}} \|\boldsymbol{J}\dot{\boldsymbol{q}} + \eta \tilde{\boldsymbol{x}}\|_2^2 \quad (10)$$
$$\text{subject to } \boldsymbol{W}\dot{\boldsymbol{q}} \preceq \boldsymbol{w},$$

in which $\boldsymbol{W} \triangleq \boldsymbol{W}(\boldsymbol{q}) \in \mathbb{R}^{r \times n}$ and $\boldsymbol{w} \triangleq \boldsymbol{w}(\boldsymbol{q}) \in \mathbb{R}^r$. Problem 10 is the optimization of a convex quadratic function since $\boldsymbol{J}^T \boldsymbol{J} \geq 0$ over the polyhedron $\boldsymbol{W}\dot{\boldsymbol{q}} \preceq \boldsymbol{w}$ [40]. Furthermore, Problem 10 does not limit the joint velocity norm; hence, it may generate unfeasible velocities. To avoid a nested optimization such as that in Problem 9, which requires a higher computational time, we resort to adding a damping factor $\lambda \in [0, \infty)$ to Problem 10 to find

$$\min_{\dot{q}} \|J\dot{q} + \eta\tilde{x}\|_2^2 + \lambda \|\dot{q}\|_2^2 \quad (11)$$

$$\text{subject to } W\dot{q} \preceq w,$$

which limits the joint velocity norm.

---

[1] Moreover, a software implementation of the proposed methodology is part of the DQ Robotics open-source robotics library (https://dqrobotics.github.io/).

[2] This paper follows the notation and terminology on convex optimization used in [40].

[3] In the optimization literature (e.g. [40]), a minimizer is usually denoted with a superscript asterisk, such as $\dot{\boldsymbol{q}}^*$. However, we use the superscript $o$, as in $\dot{\boldsymbol{q}}^o$, to avoid notational confusion with the dual-quaternion conjugate.

More than one robot can be controlled simultaneously with Problem 11 in a centralized architecture, which is suitable for most applications in surgical robotics. Suppose that $p$ robots should follow their own independent task-space trajectories. For $i = 1, \ldots, p$, let each robot $R_i$ have $n_i$ joints, a joint velocity vector $\dot{q}_i$, a task Jacobian $J_i$, and a task error $\tilde{x}_i$. Problem 10 becomes

$$\min_{\dot{g}} \|A\dot{g} + \eta\tilde{y}\|_2^2 + \lambda \|\dot{g}\|_2^2 \quad (12)$$

$$\text{subject to } M\dot{g} \preceq m,$$

in which

$$A = \begin{bmatrix} J_1 & \cdots & 0 \\ \vdots & \ddots & \vdots \\ 0 & \cdots & J_p \end{bmatrix}, \quad g = \begin{bmatrix} q_1 \\ \vdots \\ q_p \end{bmatrix}, \quad \tilde{y} = \begin{bmatrix} \tilde{x}_1 \\ \vdots \\ \tilde{x}_p \end{bmatrix},$$

$M \triangleq M(g) \in \mathbb{R}^{r \times \sum n_i}$, $m \triangleq m(g) \in \mathbb{R}^r$, and $0$ is a matrix of zeros with appropriate dimensions.

The vector field inequalities proposed in Section III use appropriate linear constraints to generate dynamic active constraints. In addition, by solving Problem 12, we locally ensure the smallest trajectory tracking error for a collision-free path.

In this work, we assume that the desired trajectory, $x_d(t)$, is generated online by the surgeon through teleoperation or comanipulation. In this case, trajectory tracking controllers that require future knowledge of the trajectory [42] cannot be used, and a set-point regulator given by the solution of Problem 12 is a proper choice.

## III. VECTOR-FIELD INEQUALITY FOR DYNAMIC ENTITIES

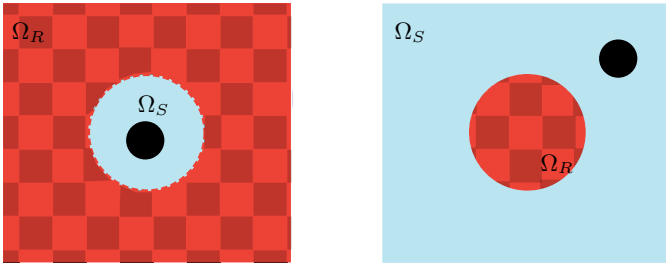

Fig. 1. The proposed vector field inequalities can have two types of behavior, in which the black circle represents a robot entity: keeping a robot entity inside a safe zone, $\Omega_S$ (left); or outside of a restricted zone, $\Omega_R$ (right). Restricted zones are checked for contrast with safe zones.

The vector-field inequality for dynamic elements requires the following:

1) A function $d \triangleq d(q, t) \in \mathbb{R}$ that encodes the (signed) distance between the two collidable entities. The *robot entity* is kinematically coupled to the robot, and the other entity, called the *restricted zone*, is part of the workspace (or part of another robot),

2) A Jacobian relating the time derivative of the distance function and the joints' velocities in the general form

$$\dot{d} = \underbrace{\frac{\partial (d(q,t))}{\partial q}}_{J_d} \dot{q} + \zeta(t), \quad (13)$$

in which the residual $\zeta(t) = \dot{d} - J_d\dot{q}$ contains the distance dynamics unrelated to the joints' velocities.

We assume that the residual is known but cannot be controlled. For instance, the residual might encode the movement of a geometrical entity that can be tracked or estimated and may be related to the workspace.[4]

Using distance functions and distance Jacobians, complex dynamic restricted zones can be generated, either by maintaining the distance above a desired level or by keeping the distance below a certain level, as shown in Fig. 1.

### A. Keeping the robot entity outside a restricted region

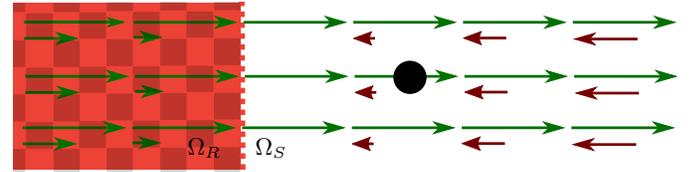

Fig. 2. A vector field inequality that keeps a point outside of the restricted zone, $\Omega_R$, whose boundary is a plane. To each point in space is assigned a maximum approach velocity (the lower vector in each vector pair), and a maximum separation velocity (the upper vector in each vector pair). The approach velocity decreases exponentially with the distance, and the maximum separating velocity is unconstrained.

To keep the robot entity outside a restricted zone, we define a minimum safe distance $d_{\text{safe}} \triangleq d_{\text{safe}}(t) \in [0, \infty)$, which delineates the time-dependent boundary of the restricted zone, and a signed distance

$$\tilde{d} \triangleq \tilde{d}(q, t) = d - d_{\text{safe}}. \quad (14)$$

The restricted zone $\Omega_R$ and safe zone $\Omega_S$ are

$$\Omega_R \triangleq \left\{ q \in \mathbb{R}^n, t \in [0, \infty) : \tilde{d}(q, t) < 0 \right\},$$

$$\Omega_S \triangleq \left\{ q \in \mathbb{R}^n, t \in [0, \infty) : \tilde{d}(q, t) \geq 0 \right\}.$$

The signed distance dynamics is given by

$$\dot{\tilde{d}} = \dot{d} - \dot{d}_{\text{safe}}. \quad (15)$$

A positive $\dot{\tilde{d}}$ means that the robot entity and restricted zone are moving away from each other, whereas a negative $\dot{\tilde{d}}$ means that they are moving closer to each other.

Given $\eta_d \in [0, \infty)$, the signed distance dynamics is constrained by [30]

$$\dot{\tilde{d}} \geq -\eta_d \tilde{d}. \quad (16)$$

Constraint 16 assigns a velocity constraint for the robot entity to each point in space, as shown in Fig. 2, which has at least exponential behavior according to Gronwall's lemma [31]. To understand the physical meaning of the constraint, first suppose that $\tilde{d}(\boldsymbol{q}, 0) \geq 0$, which means that the robot entity is in the safe zone when $t = 0$. In this situation, *any* increase in the distance is *always* allowed, which implies that $\dot{\tilde{d}} \geq 0 \geq -\eta_d \tilde{d}$. However, when the distance decreases, $0 \geq \dot{\tilde{d}} \geq -\eta_d \tilde{d}$ and the *maximum* rate of decrease in the distance rate is exponential, given by $\dot{\tilde{d}} = -\eta_d \tilde{d}$. In this way, as the robot entity more closely approaches the restricted zone, the allowed approach velocity between the restricted zone and the system gets smaller. Any slower approaching motion is also allowed; hence, $\dot{\tilde{d}} \geq -\eta_d \tilde{d}$. As soon as $\tilde{d} = 0$, the restriction becomes $\dot{\tilde{d}} \geq 0$; therefore, the robot entity will not enter the restricted zone.

---

[4]The vector-field inequalities proposed in our earlier work [32] are a special case of the framework proposed in this work with $\zeta(t) = 0 \, \forall t$.

Now consider that $\tilde{d}(\boldsymbol{q}, 0) < 0$; that is, the system starts inside the restricted zone. In this case, Constraint 16 will only be fulfilled if $\dot{\tilde{d}} \geq \eta_d \left| \tilde{d} \right|$, which means that the system will, at least, be pushed towards the safe zone with *minimum* rate of decrease in the distance given by $\dot{\tilde{d}} = -\eta_d \tilde{d} = \eta_d \left| \tilde{d} \right|$.

Using (13) and (15), Constraint 16 is written explicitly in terms of the joint velocities as

$$\boldsymbol{J}_d \dot{\boldsymbol{q}} \geq -\eta_d \tilde{d} - \zeta_{\text{safe}}(t), \quad (17)$$

where the residual $\zeta_{\text{safe}}(t) \triangleq \zeta(t) - \dot{d}_{\text{safe}}$ takes into account the effects, on the distance, of a moving obstacle with residual $\zeta(t)$ and a time-varying safe-zone boundary $\dot{d}_{\text{safe}}$. If $\zeta_{\text{safe}} > 0$, the restricted zone contributes[5] to an increase in the distance between itself and the robot entity. Conversely, if $\zeta_{\text{safe}} < 0$, then the restricted zone contributes to a decrease in the distance between itself and the robot entity. If $\zeta_{\text{safe}} + \eta_d \tilde{d} < 0$, then the robot has to actively move away from the restricted zone.

To fit into Problem 12, we rewrite Constraint 17 as

$$-\boldsymbol{J}_d \dot{\boldsymbol{q}} \leq \eta_d \tilde{d} + \zeta_{\text{safe}}(t). \quad (18)$$

Note that any number of constraints in the form of Constraint 18 can be found for different interactions between robot entities and restricted zones in the robot workspace. Moreover, by describing the interaction as a distance function, complex interactions will be only one-degree-of-freedom (one DOF) constraints.

*Remark* 1. When the distance is obtained using the Euclidean norm, its derivative is singular when the norm is zero. The squared distance is useful to avoid such singularities, as shown in Sections IV-C and IV-E. When the squared distance $D \triangleq d^2$ is used, the signed distance in (14) is redefined as $\tilde{D} \triangleq \tilde{D}(\boldsymbol{q}, t) = D - D_{\text{safe}}$, in which $D_{\text{safe}} \triangleq d_{\text{safe}}^2$ [39]. Constraint 16 then becomes $\dot{\tilde{D}} \geq -\eta_D \tilde{D}$, and any reasoning is otherwise unaltered.

### B. Keeping the robot entity inside a safe region

Using the same methodology of Section III-A, we redefine $d_{\text{safe}}$ to maintain the robot *inside* a safe region; that is,

$$\tilde{d} \triangleq d_{\text{safe}} - d,$$

[5] The motion of the restricted zone takes into account the actual obstacle motion and the motion of the safe boundary (i.e., when $\dot{d}_{\text{safe}} \neq 0$).

with final solution, assuming the desired signed distance dynamics (16), given by

$$\boldsymbol{J}_d \dot{\boldsymbol{q}} \leq \eta_d \tilde{d} - \zeta_{\text{safe}}. \quad (19)$$

## IV. SQUARED DISTANCE FUNCTIONS AND CORRESPONDING JACOBIANS

In order to use the vector-field inequalities in (18) and (19), we define the (squared) distance functions for relevant geometrical primitives (point, line, and plane); then, we find the corresponding Jacobians and residuals. These geometrical primitives can be easily combined to obtain other primitives. For instance, a point combined with a positive scalar yields a sphere, whereas a line combined with a positive scalar yields an infinite cylinder. The intersection between an infinite cylinder and two parallel planes results in a finite cylinder. Polyhedra can be defined as intersections of planes. Table II summarizes the distance functions and corresponding Jacobians.

TABLE II
SUMMARY OF PRIMITIVES

| Primitive | Distance function | | Jacobian | | Residual | |
|---|---|---|---|---|---|---|
| Point-to-Point | $D_{\boldsymbol{t},\boldsymbol{p}}$ | (Eq. 21) | $\boldsymbol{J}_{\boldsymbol{t},\boldsymbol{p}}$ | (Eq. 22) | $\zeta_{\boldsymbol{t},\boldsymbol{p}}$ | (Eq. 22) |
| Point-to-Line | $D_{\boldsymbol{t},\boldsymbol{l}}$ | (Eq. 29) | $\boldsymbol{J}_{\boldsymbol{t},\boldsymbol{l}}$ | (Eq. 32) | $\zeta_{\boldsymbol{t},\boldsymbol{l}}$ | (Eq. 32) |
| Line-to-Point | $D_{\boldsymbol{l}_z,\boldsymbol{p}}$ | (Eq. 33) | $\boldsymbol{J}_{\boldsymbol{l}_z,\boldsymbol{p}}$ | (Eq. 34) | $\zeta_{\boldsymbol{l}_z,\boldsymbol{p}}$ | (Eq. 34) |
| Line-to-Line | $D_{\boldsymbol{l}_z,\boldsymbol{l}}$ | (Eq. 50) | $\boldsymbol{J}_{\boldsymbol{l}_z,\boldsymbol{l}}$ | (Eq. 48) | $\zeta_{\boldsymbol{l}_z,\boldsymbol{l}}$ | (Eq. 49) |
| Plane-to-Point | $d^{\pi_z}_{\pi_z,\boldsymbol{p}}$ | (Eq. 54) | $\boldsymbol{J}_{\pi_z,\boldsymbol{p}}$ | (Eq. 56) | $\zeta_{\pi_z,\boldsymbol{p}}$ | (Eq. 55) |
| Point-to-Plane | $d^{\pi}_{\boldsymbol{t},\pi}$ | (Eq. 57) | $\boldsymbol{J}_{\boldsymbol{t},\pi}$ | (Eq. 59) | $\zeta_{\pi}$ | (Eq. 58) |

### A. Point-to-point squared distance $D_{\boldsymbol{t},\boldsymbol{p}}$ and Jacobian $\boldsymbol{J}_{\boldsymbol{t},\boldsymbol{p}}$ of the manipulator

The Euclidean distance between two points $\boldsymbol{p}_1, \boldsymbol{p}_2 \in \mathbb{H}_p$ is given by[6]

$$d_{\boldsymbol{p}_1,\boldsymbol{p}_2} = \| \boldsymbol{p}_1 - \boldsymbol{p}_2 \|. \quad (20)$$

Since the time derivative of (20) is singular at $d = 0$ [39], we use the squared distance, whose time derivative is defined everywhere:

$$D_{\boldsymbol{p}_1,\boldsymbol{p}_2} \triangleq d^2_{\boldsymbol{p}_1,\boldsymbol{p}_2} = \| \boldsymbol{p}_1 - \boldsymbol{p}_2 \|^2. \quad (21)$$

Given a point $\boldsymbol{t} \triangleq \boldsymbol{t}(\boldsymbol{q}(t)) \in \mathbb{H}_p$ in the robot, where $\boldsymbol{q}(t) \in \mathbb{R}^n$ is the time-varying joints' configuration, and an arbitrary point $\boldsymbol{p} \triangleq \boldsymbol{p}(t) \in \mathbb{H}_p$ in space, we use (4) and (6) to find that

$$\frac{d}{dt}(D_{\boldsymbol{t},\boldsymbol{p}}) = \underbrace{2\operatorname{vec}_4(\boldsymbol{t}-\boldsymbol{p})^T \boldsymbol{J}_{\boldsymbol{t}}}_{\boldsymbol{J}_{\boldsymbol{t},\boldsymbol{p}}} \dot{\boldsymbol{q}} + \underbrace{2\langle \boldsymbol{t}-\boldsymbol{p}, -\dot{\boldsymbol{p}} \rangle}_{\zeta_{\boldsymbol{t},\boldsymbol{p}}}. \quad (22)$$

### B. Dual-quaternion line $\underline{\boldsymbol{l}}_z$ and line Jacobian $\boldsymbol{J}_{\underline{l}_z}$

In minimally invasive surgery and microsurgery, tools have long and thin shafts, which may have some parts outside the field of view; therefore, collisions with other tools or the environment might happen. By describing tool shafts using Plücker lines, such interactions can be mathematically modeled. A Plücker line (see Fig. 3) belongs to the set $\mathcal{H}_p \cap \underline{\boldsymbol{S}}$ and thus is represented by a unit dual quaternion such as [34], [37]

$$\underline{\boldsymbol{l}} = \boldsymbol{l} + \varepsilon \boldsymbol{m}, \quad (23)$$

[6] The quaternion norm is equivalent to the Euclidean norm.

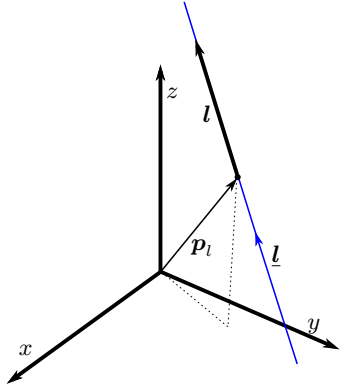

Fig. 3. Representation of a Plücker line $\underline{\boldsymbol{l}} \triangleq \boldsymbol{l} + \varepsilon \boldsymbol{m}$, where $\boldsymbol{l}$ is the line direction, $\boldsymbol{m} = \boldsymbol{p}_l \times \boldsymbol{l}$ is the line moment, and $\boldsymbol{p}_l$ is an arbitrary point in the line.

where $\boldsymbol{l} \in \mathbb{H}_p \cap \mathbb{S}^3$ is a pure quaternion with a unit norm that represents the line direction and the line moment is given by $\boldsymbol{m} = \boldsymbol{p}_l \times \boldsymbol{l}$, in which $\boldsymbol{p}_l \in \mathbb{H}_p$ is an arbitrary point on the line.

When the forward kinematics of the robot is described using the Denavit–Hartenberg convention, the $z$-axis of each frame passes through the corresponding joint actuation axis [11].[7] Hence, it is especially useful to find the Jacobian relating the joint velocities and the line that passes through the $z$-axis of a reference frame in the robot.

Consider a Plücker line on top of the $z$-axis of a frame of interest, whose pose is given by $\underline{\boldsymbol{x}} = \boldsymbol{r} + \frac{1}{2}\varepsilon \boldsymbol{t}\boldsymbol{r}$, in a robotic manipulator kinematic chain. Since the frame of interest changes according to the robot motion, the line changes accordingly; hence,

$$\underline{\boldsymbol{l}}_z \triangleq \underline{\boldsymbol{l}}_z\left(\boldsymbol{q}\left(t\right)\right) = \boldsymbol{l}_z + \varepsilon \boldsymbol{m}_z, \quad (24)$$

where $\boldsymbol{l}_z = \boldsymbol{r}\hat{k}\boldsymbol{r}^*$ and $\boldsymbol{m}_z = \boldsymbol{t} \times \boldsymbol{l}_z$. The derivative of (24) is

$$\underline{\dot{\boldsymbol{l}}}_z = \dot{\boldsymbol{l}}_z + \varepsilon \dot{\boldsymbol{m}}_z, \quad (25)$$

where $\dot{\boldsymbol{l}}_z = \dot{\boldsymbol{r}}\hat{k}\boldsymbol{r}^* + \boldsymbol{r}\hat{k}\dot{\boldsymbol{r}}^*$; thus,

$$\mathrm{vec}_4 \dot{\boldsymbol{l}}_z = \underbrace{\left(\overline{\boldsymbol{H}}_4\left(\hat{k}\boldsymbol{r}^*\right)\boldsymbol{J}_r + \overset{+}{\boldsymbol{H}}_4\left(\boldsymbol{r}\hat{k}\right)\boldsymbol{C}_4\boldsymbol{J}_r\right)}_{\boldsymbol{J}_{r_z}}\dot{\boldsymbol{q}}, \quad (26)$$

in which $\boldsymbol{C}_4 = \mathrm{diag}(1,-1,-1,-1)$ and $\mathrm{vec}_4 \dot{\boldsymbol{r}}$ is given by (7). In addition, the derivative of the line moment is $\dot{\boldsymbol{m}}_z = \dot{\boldsymbol{t}} \times \boldsymbol{l}_z + \boldsymbol{t} \times \dot{\boldsymbol{l}}_z$; hence, we use (3), (6), and (26) to obtain

$$\mathrm{vec}_4 \dot{\boldsymbol{m}}_z = \underbrace{\left(\overline{\boldsymbol{S}}\left(\boldsymbol{l}_z\right)^T \boldsymbol{J}_t + \overline{\boldsymbol{S}}\left(\boldsymbol{t}\right)\boldsymbol{J}_{r_z}\right)}_{\boldsymbol{J}_{m_z}}\dot{\boldsymbol{q}}. \quad (27)$$

Finally, we rewrite (25) explicitly in term of the joint velocities using (26) and (27) as

$$\mathrm{vec}_8 \underline{\dot{\boldsymbol{l}}}_z = \begin{bmatrix} \boldsymbol{J}_{r_z} \\ \boldsymbol{J}_{m_z} \end{bmatrix}\dot{\boldsymbol{q}} \triangleq \boldsymbol{J}_{l_z}\dot{\boldsymbol{q}}. \quad (28)$$

[7]The method described here is sufficiently general and different conventions can be used (for instance, the joint actuation axis could be the $x$- or $y$-axis), such that the derivation of the corresponding equations follows the same procedure.

*C. Point-to-line squared distance $D_{t,l}$ and Jacobian of the manipulator $\boldsymbol{J}_{t,l}$*

Considering the robot joints' configuration vector $\boldsymbol{q}(t) \in \mathbb{R}^n$ and a point $\boldsymbol{t} \triangleq \boldsymbol{t}(\boldsymbol{q}(t)) \in \mathbb{H}_p$ in the robot, the squared distance between $\boldsymbol{t}$ and a arbitrary line $\underline{\boldsymbol{l}} \triangleq \underline{\boldsymbol{l}}(t) \in \mathcal{H}_p \cap \underline{\mathcal{S}}$ in space is given by [43]

$$D_{t,l} \triangleq \|\boldsymbol{t} \times \boldsymbol{l} - \boldsymbol{m}\|^2. \quad (29)$$

For notational convenience, we define $\boldsymbol{h}_{A1} \triangleq \boldsymbol{t} \times \boldsymbol{l} - \boldsymbol{m}$, whose derivative is

$$\dot{\boldsymbol{h}}_{A1} = \dot{\boldsymbol{t}} \times \boldsymbol{l} + \underbrace{\boldsymbol{t} \times \dot{\boldsymbol{l}} - \dot{\boldsymbol{m}}}_{\boldsymbol{h}_{A2}}. \quad (30)$$

Using (2), (3), (4), and 30, we obtain

$$\begin{aligned}\dot{D}_{t,l} &= 2\langle \dot{\boldsymbol{h}}_{A1}, \boldsymbol{h}_{A1}\rangle \\ &= 2\langle(\dot{\boldsymbol{t}} \times \boldsymbol{l}), \boldsymbol{h}_{A1}\rangle + 2\langle \boldsymbol{h}_{A2}, \boldsymbol{h}_{A1}\rangle \\ &= 2\,\mathrm{vec}_4\left(\boldsymbol{h}_{A1}\right)^T \overline{\boldsymbol{S}}\left(\boldsymbol{l}\right)^T \mathrm{vec}_4\left(\dot{\boldsymbol{t}}\right) + 2\langle \boldsymbol{h}_{A2}, \boldsymbol{h}_{A1}\rangle.\end{aligned} \quad (31)$$

Letting $\zeta_{t,l} \triangleq 2\langle \boldsymbol{h}_{A2}, \boldsymbol{h}_{A1}\rangle$, we use (6) to obtain the squared-distance time derivative explicitly in terms of the robot joints' velocities:

$$\dot{D}_{t,l} = \underbrace{2\,\mathrm{vec}_4\left(\boldsymbol{t} \times \boldsymbol{l} - \boldsymbol{m}\right)^T \overline{\boldsymbol{S}}\left(\boldsymbol{l}\right)^T \boldsymbol{J}_t}_{\boldsymbol{J}_{t,l}}\dot{\boldsymbol{q}} + \zeta_{t,l}. \quad (32)$$

*D. Line-to-point squared distance $D_{l_z,p}$ and Jacobian $\boldsymbol{J}_{l_z,p}$ of the manipulator*

Given the line $\underline{\boldsymbol{l}}_z$, as in (24), on top of the $z$-axis of a frame rigidly attached to the robot and an arbitrary point $\boldsymbol{p} \in \mathbb{H}_p$ in space, we use (29) to obtain

$$D_{l_z,p} = \|\boldsymbol{p} \times \boldsymbol{l}_z - \boldsymbol{m}_z\|^2. \quad (33)$$

Similar to (32), we write the time derivative of (33) explicitly in terms of the joints' velocities:

$$\dot{D}_{l_z,p} = \underbrace{2\,\mathrm{vec}_4\left(\boldsymbol{p} \times \boldsymbol{l}_z - \boldsymbol{m}_z\right)^T \left(\overline{\boldsymbol{S}}\left(\boldsymbol{p}\right)\boldsymbol{J}_{r_z} - \boldsymbol{J}_{m_z}\right)}_{\boldsymbol{J}_{l_z,p}}\dot{\boldsymbol{q}}$$
$$+ \underbrace{2\langle \dot{\boldsymbol{p}} \times \boldsymbol{l}_z, \boldsymbol{p} \times \boldsymbol{l}_z - \boldsymbol{m}_z\rangle}_{\zeta_{l_z,p}}. \quad (34)$$

This primitive (i.e., the function $D_{l_z,p}$ together with its derivative $\dot{D}_{l_z,p}$) is useful to define a compliant remote center of motion by having a nonzero safe distance.

*E. Line-to-line distance $D_{l_z,l}$ and Jacobian $\boldsymbol{J}_{l_z,l}$ of the manipulator*

The distance between two Plücker lines $\underline{\boldsymbol{l}}_1, \underline{\boldsymbol{l}}_2 \in \mathcal{H}_p \cap \underline{\mathcal{S}}$ is obtained through the inner-product and cross-product operations. The inner product between $\underline{\boldsymbol{l}}_1$ and $\underline{\boldsymbol{l}}_2$ is [34][8]

$$\begin{aligned}\langle \underline{\boldsymbol{l}}_1, \underline{\boldsymbol{l}}_2\rangle &= \|\underline{\boldsymbol{l}}_1\|\|\underline{\boldsymbol{l}}_2\|\cos\left(\phi_{1,2} + \varepsilon d_{1,2}\right) \\ &= \cos\phi_{1,2} - \varepsilon\sin\phi_{1,2}d_{1,2},\end{aligned} \quad (35)$$

[8]Given a function $f : \mathbb{D} \to \mathbb{D}$, where $\mathbb{D} \triangleq \{\underline{\boldsymbol{h}} \in \mathcal{H} : \mathrm{Im}\left(\underline{\boldsymbol{h}}\right) = 0\}$, it is possible to show that $f\left(a + \varepsilon b\right) = f\left(a\right) + \varepsilon b f'\left(a\right)$. For more details, see [37].

where $d_{1,2} \in [0, \infty)$ and $\phi_{1,2} \in [0, 2\pi)$ are the distance and the angle between $\underline{l}_1$ and $\underline{l}_2$, respectively. Moreover, given $\underline{s}_{1,2} \in \mathcal{H}_p \cap \underline{\mathcal{S}}$—the line perpendicular to both $\underline{l}_1$ and $\underline{l}_2$—such that $\underline{s}_{1,2} \triangleq s_{1,2} + \varepsilon m_{s_{1,2}}$, the cross product between $\underline{l}_1$ and $\underline{l}_2$ is [34]

$$\begin{aligned} \underline{l}_1 \times \underline{l}_2 &= \|\underline{l}_1\| \|\underline{l}_2\| \, \underline{s}_{1,2} \sin(\phi_{1,2} + \varepsilon d_{1,2}) \\ &= (s_{1,2} + \varepsilon m_{s_{1,2}})(\sin\phi_{1,2} + \varepsilon \cos\phi_{1,2} d_{1,2}) \\ &= s_{1,2} \sin\phi_{1,2} + \varepsilon \left( m_{s_{1,2}} \sin\phi_{1,2} + s_{1,2} \cos\phi_{1,2} d_{1,2} \right). \end{aligned} \quad (36)$$

The squared distance between $\underline{l}_1$ and $\underline{l}_2$ when they are not parallel (i.e., $\phi_{1,2} \in (0, 2\pi) \setminus \pi$) is obtained using (35) and (36):

$$D_{1,2\nparallel} = \frac{\|\mathcal{D}(\langle \underline{l}_1, \underline{l}_2 \rangle)\|^2}{\|\mathcal{P}(\underline{l}_1 \times \underline{l}_2)\|^2} = \frac{\|d_{1,2} \sin\phi_{1,2}\|^2}{\|s_{1,2} \sin\phi_{1,2}\|^2} = d_{1,2}^2, \quad (37)$$

The squared distance between $\underline{l}_1$ and $\underline{l}_2$ when they are parallel (i.e., $\phi_{1,2} \in \{0, \pi\}$) is obtained as

$$D_{1,2\parallel} \triangleq \|\mathcal{D}(\underline{l}_1 \times \underline{l}_2)\|^2 = \|s_{1,2} d_{1,2}\|^2 = d_{1,2}^2. \quad (38)$$

To find the distance Jacobian and residual between a line $\underline{l}_z$ in the robot and an arbitrary line $\underline{l}$ in space, we begin by finding the inner-product Jacobian and residual in Section IV-E1 and the cross product Jacobian and residual in Section IV-E2. Those are used in Section IV-E3 to find the derivative of (37), and in Section IV-E4 to find the derivative of (38). Finally, they are unified in the final form of the distance Jacobian and residual in Section IV-E5.

*1) Inner-product Jacobian, $J_{\langle \underline{l}_z, \underline{l} \rangle}$:* The time derivative of the inner product between $\underline{l}_z, \underline{l} \in \mathcal{H}_p \cap \underline{\mathcal{S}}$ is given by

$$\frac{d}{dt}(\langle \underline{l}_z, \underline{l} \rangle) = \langle \underline{\dot{l}}_z, \underline{l} \rangle + \langle \underline{l}_z, \underline{\dot{l}} \rangle = -\frac{1}{2} \left( \underline{\dot{l}}_z \underline{l} + \underline{l} \underline{\dot{l}}_z \right) + \underbrace{\langle \underline{l}_z, \underline{\dot{l}} \rangle}_{\underline{\zeta}_{\langle \underline{l}_z, \underline{l} \rangle}}.$$

Hence, using (28) we obtain

$$\mathrm{vec}_8 \frac{d}{dt}(\langle \underline{l}_z, \underline{l} \rangle) = \overbrace{-\frac{1}{2} \left( \overline{\boldsymbol{H}}_8(\underline{l}) + \overset{+}{\boldsymbol{H}}_8(\underline{l}) \right) \boldsymbol{J}_{l_z}}^{\boldsymbol{J}_{\langle \underline{l}_z, \underline{l} \rangle}} \dot{\boldsymbol{q}} + \mathrm{vec}_8 \underline{\zeta}_{\langle \underline{l}_z, \underline{l} \rangle},$$

which can be explicitly written in terms of the primary and dual parts as

$$\begin{bmatrix} \mathrm{vec}_4 \dot{\mathcal{P}}(\langle \underline{l}_z, \underline{l} \rangle) \\ \mathrm{vec}_4 \dot{\mathcal{D}}(\langle \underline{l}_z, \underline{l} \rangle) \end{bmatrix} = \begin{bmatrix} \boldsymbol{J}_{\mathcal{P}(\langle \underline{l}_z, \underline{l} \rangle)} \\ \boldsymbol{J}_{\mathcal{D}(\langle \underline{l}_z, \underline{l} \rangle)} \end{bmatrix} \dot{\boldsymbol{q}} + \begin{bmatrix} \mathrm{vec}_4 \mathcal{P}(\underline{\zeta}_{\langle \underline{l}_z, \underline{l} \rangle}) \\ \mathrm{vec}_4 \mathcal{D}(\underline{\zeta}_{\langle \underline{l}_z, \underline{l} \rangle}) \end{bmatrix}. \quad (39)$$

*2) Cross-product Jacobian, $J_{\underline{l}_z \times \underline{l}}$:* The time derivative of the cross product between $\underline{l}_z, \underline{l} \in \mathcal{H}_p \cap \underline{\mathcal{S}}$ is given by

$$\frac{d}{dt}(\underline{l}_z \times \underline{l}) = \underline{\dot{l}}_z \times \underline{l} + \underline{l}_z \times \underline{\dot{l}} = \frac{1}{2} \left( \underline{\dot{l}}_z \underline{l} - \underline{l} \underline{\dot{l}}_z \right) + \underbrace{\underline{l}_z \times \underline{\dot{l}}}_{\underline{\zeta}_{\underline{l}_z \times \underline{l}}}.$$

Using (28), we obtain

$$\mathrm{vec}_8 \frac{d}{dt}(\underline{l}_z \times \underline{l}) = \overbrace{\frac{1}{2} \left( \overline{\boldsymbol{H}}_8(\underline{l}) - \overset{+}{\boldsymbol{H}}_8(\underline{l}) \right) \boldsymbol{J}_{l_z}}^{\boldsymbol{J}_{\underline{l}_z \times \underline{l}}} \dot{\boldsymbol{q}} + \mathrm{vec}_8 \underline{\zeta}_{\underline{l}_z \times \underline{l}},$$

which can be explicitly written in terms of the primary and dual parts as

$$\begin{bmatrix} \mathrm{vec}_4 \dot{\mathcal{P}}(\underline{l}_z \times \underline{l}) \\ \mathrm{vec}_4 \dot{\mathcal{D}}(\underline{l}_z \times \underline{l}) \end{bmatrix} = \begin{bmatrix} \boldsymbol{J}_{\mathcal{P}(\underline{l}_z \times \underline{l})} \\ \boldsymbol{J}_{\mathcal{D}(\underline{l}_z \times \underline{l})} \end{bmatrix} \dot{\boldsymbol{q}} + \begin{bmatrix} \mathrm{vec}_4 \mathcal{P}(\underline{\zeta}_{\underline{l}_z \times \underline{l}}) \\ \mathrm{vec}_4 \mathcal{D}(\underline{\zeta}_{\underline{l}_z \times \underline{l}}) \end{bmatrix} \quad (40)$$

*3) Nonparallel distance Jacobian, $J_{\underline{l}_z, \underline{l}\nparallel}$:* From (37), the squared distance between two nonparallel lines $\underline{l}_z, \underline{l} \in \mathcal{H}_p \cap \underline{\mathcal{S}}$ is

$$D_{\underline{l}_z, \underline{l}\nparallel} = \frac{\|\mathcal{D}(\langle \underline{l}_z, \underline{l} \rangle)\|^2}{\|\mathcal{P}(\underline{l}_z \times \underline{l})\|^2}, \quad (41)$$

with time derivative given by

$$\dot{D}_{\underline{l}_z, \underline{l}\nparallel} = \overbrace{\frac{1}{\|\mathcal{P}(\underline{l}_z \times \underline{l})\|^2}}^{a} \frac{d}{dt} \left( \|\mathcal{D}(\langle \underline{l}_z, \underline{l} \rangle)\|^2 \right) \\ -\underbrace{\frac{\|\mathcal{D}(\langle \underline{l}_z, \underline{l} \rangle)\|^2}{\|\mathcal{P}(\underline{l}_z \times \underline{l})\|^4}}_{b} \frac{d}{dt} \left( \|\mathcal{P}(\underline{l}_z \times \underline{l})\|^2 \right). \quad (42)$$

We use (4) and (39) to obtain

$$\frac{d}{dt} \left( \|\mathcal{D}(\langle \underline{l}_z, \underline{l} \rangle)\|^2 \right) = 2 \overbrace{\mathrm{vec}_4 \mathcal{D}(\langle \underline{l}_z, \underline{l} \rangle)^T \boldsymbol{J}_{\mathcal{D}(\langle \underline{l}_z, \underline{l} \rangle)}}^{\boldsymbol{J}_{\|\mathcal{D}(\langle \underline{l}_z, \underline{l} \rangle)\|^2}} \dot{\boldsymbol{q}} \\ + \underbrace{2 \mathrm{vec}_4 \mathcal{D}(\langle \underline{l}_z, \underline{l} \rangle)^T \mathrm{vec}_4 \mathcal{D}(\underline{\zeta}_{\langle \underline{l}_z, \underline{l} \rangle})}_{\zeta_{\|\mathcal{D}(\langle \underline{l}_z, \underline{l} \rangle)\|^2}}. \quad (43)$$

Similarly, we use (4) and (40) to obtain

$$\frac{d}{dt} \left( \|\mathcal{P}(\underline{l}_z \times \underline{l})\|^2 \right) = 2 \overbrace{\mathrm{vec}_4 \mathcal{P}(\underline{l}_z \times \underline{l})^T \boldsymbol{J}_{\mathcal{P}(\underline{l}_z \times \underline{l})}}^{\boldsymbol{J}_{\|\mathcal{P}(\underline{l}_z \times \underline{l})\|^2}} \dot{\boldsymbol{q}} \\ + \underbrace{2 \mathrm{vec}_4 \mathcal{P}(\underline{l}_z \times \underline{l})^T \mathrm{vec}_4 \mathcal{P}(\underline{\zeta}_{\underline{l}_z \times \underline{l}})}_{\zeta_{\|\mathcal{P}(\underline{l}_z \times \underline{l})\|^2}}. \quad (44)$$

Finally, replacing (43) and (44) in (42) yields

$$\dot{D}_{\underline{l}_z, \underline{l}\nparallel} = \overbrace{\left( a \boldsymbol{J}_{\|\mathcal{D}(\langle \underline{l}_z, \underline{l} \rangle)\|^2} + b \boldsymbol{J}_{\|\mathcal{P}(\underline{l}_z \times \underline{l})\|^2} \right)}^{\boldsymbol{J}_{\underline{l}_z, \underline{l}\nparallel}} \dot{\boldsymbol{q}} \\ + \underbrace{a \zeta_{\|\mathcal{D}(\langle \underline{l}_z, \underline{l} \rangle)\|^2} + b \zeta_{\|\mathcal{P}(\underline{l}_z \times \underline{l})\|^2}}_{\zeta_{\underline{l}_z, \underline{l}\nparallel}}. \quad (45)$$

*4) Parallel distance Jacobian, $J_{\underline{l}_z, \underline{l}\parallel}$:* In the degenerate case where $\underline{l}_z$ and $\underline{l}$ are parallel, the squared distance $D_{\underline{l}_z, \underline{l}\parallel}$ is given by (38); that is,

$$D_{\underline{l}_z, \underline{l}\parallel} = \|\mathcal{D}(\underline{l}_z \times \underline{l})\|^2. \quad (46)$$

Using (4) and (40), the derivative of (46) is given by

$$\dot{D}_{\underline{l}_z, \underline{l}\parallel} = 2 \overbrace{\mathrm{vec}_4 (\mathcal{D}(\underline{l}_z \times \underline{l}))^T \boldsymbol{J}_{\mathcal{D}(\underline{l}_z \times \underline{l})}}^{\boldsymbol{J}_{\underline{l}_z, \underline{l}\parallel}} \dot{\boldsymbol{q}} \\ + \underbrace{2 \mathrm{vec}_4 (\mathcal{D}(\underline{l}_z \times \underline{l}))^T \mathrm{vec}_4 \mathcal{D}(\underline{\zeta}_{\underline{l}_z \times \underline{l}})}_{\zeta_{\underline{l}_z, \underline{l}\parallel}}. \quad (47)$$

*5) Line-to-line distance Jacobian, $J_{l_z,l}$ of the manipulator:* The line-to-line distance Jacobian and residual are given by taking into account both (45) and (47) as

$$J_{l_z,l} = \begin{cases} J_{l_z,l \not\parallel} & \phi \in (0, 2\pi) \setminus \pi \\ J_{l_z,l \parallel} & \phi \in \{0, \pi\} \end{cases}, \quad (48)$$

$$\zeta_{l_z,l} = \begin{cases} \zeta_{l_z,l \not\parallel} & \phi \in (0, 2\pi) \setminus \pi \\ \zeta_{l_z,l \parallel} & \phi \in \{0, \pi\} \end{cases}, \quad (49)$$

in which $\phi = \arccos \mathcal{P}(\langle \underline{l}, \underline{l}_z \rangle)$.

Similarly, the distance between lines is given by taking into account both (41) and (46) as

$$D_{l_z,l} = \begin{cases} D_{l_z,l \not\parallel} & \phi \in (0, 2\pi) \setminus \pi \\ D_{l_z,l \parallel} & \phi \in \{0, \pi\} \end{cases}. \quad (50)$$

*F. Plane-to-point distance $d^{\pi_z}_{\pi_z,p}$ and Jacobian $J_{\pi_z,p}$ of the manipulator*

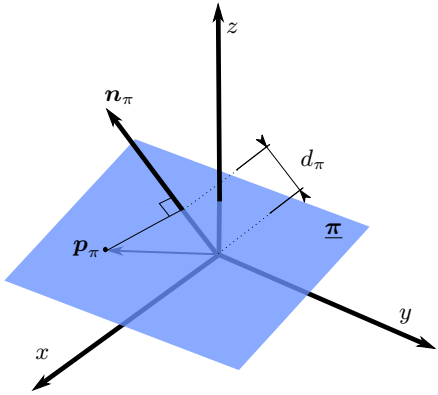

Fig. 4. Representation of a plane $\underline{\pi} \triangleq n_\pi + \varepsilon d_\pi$ in dual quaternion space. The pure quaternion $n_\pi$ is the normal to the plane and $d_\pi = \langle p_\pi, n_\pi \rangle$ is the (signed) distance between the plane and the origin of the reference frame, where $p_\pi$ is an arbitrary point in the plane.

To obtain the required relations between a plane $\underline{\pi}$ and a point, we use the plane representation in dual quaternion space, which is given by [37]

$$\underline{\pi} \triangleq n_\pi + \varepsilon d_\pi, \quad (51)$$

where $n_\pi \in \mathbb{H}_p \cap \mathbb{S}^3$ is the normal to the plane and $d_\pi \in \mathbb{R}$ is the signed perpendicular distance between the plane and the origin of the reference frame. Moreover, given an arbitrary point $p_\pi$ in the plane, the signed perpendicular distance is given by $d_\pi = \langle p_\pi, n_\pi \rangle$, as shown in Fig. 4.

Consider a frame in the robot kinematic chain with a pose given by $\underline{x} = (r + \varepsilon(1/2)tr) \in \underline{\mathcal{S}}$ and a plane $\underline{\pi}_z$, which passes through the origin of that frame, whose normal is in the same direction of the frame's $z$-axis. Since $\underline{x}$ changes according to the robot configuration, i.e., $\underline{x} \triangleq \underline{x}(q(t))$, the plane also depends on the robot configuration; that is,

$$\underline{\pi}_z \triangleq \underline{\pi}_z(q(t)) = n_{\pi_z} + \varepsilon d_{\pi_z}, \quad (52)$$

where $n_{\pi_z} = r\hat{k}r^*$ and $d_{\pi_z} = \langle t, n_{\pi_z} \rangle$. Since $\dot{d}_{\pi_z} = \langle \dot{t}, n_{\pi_z} \rangle + \langle t, \dot{n}_{\pi_z} \rangle$, we use (2), (6), and (26) to obtain

$$\dot{d}_{\pi_z} = \underbrace{\left( \mathrm{vec}_4(n_{\pi_z})^T J_t + \mathrm{vec}_4(t)^T J_{r_z} \right)}_{J_{d_{\pi_z}}} \dot{q}. \quad (53)$$

Furthermore, as the signed distance between an arbitrary point $p \in \mathbb{H}_p$ in space and the plane $\underline{\pi}_z$ from the point of view of the plane [32] is

$$d^{\pi_z}_{\pi_z,p} = \langle p, n_{\pi_z} \rangle - d_{\pi_z}, \quad (54)$$

we use (2), (26) and (53) to obtain

$$\dot{d}^{\pi_z}_{\pi_z,p} = \langle p, \dot{n}_{\pi_z} \rangle - \dot{d}_{\pi_z} + \overbrace{\langle \dot{p}, n_{\pi_z} \rangle}^{\zeta_{\pi_z,p}} \quad (55)$$

$$= \underbrace{\left( \mathrm{vec}_4(p)^T J_{r_z} - J_{d_{\pi_z}} \right)}_{J_{\pi_z,p}} \dot{q} + \zeta_{\pi_z,p}. \quad (56)$$

*G. Point-to-plane distance $d^\pi_{t,\pi}$ and Jacobian $J_{t,\pi}$ of the manipulator*

If $\mathcal{F}_\pi$ is a frame attached to an arbitrary plane in space, the signed distance between a point $t \triangleq t(q(t)) \in \mathbb{H}_p$ in the robot and the plane $\underline{\pi}$ from the point of view of the plane is given by

$$d^\pi_{t,\pi} = \langle t, n_\pi \rangle - d_\pi. \quad (57)$$

Using (2) and (6), the time derivative of (57) is

$$\dot{d}^\pi_{t,\pi} = \langle \dot{t}, n_\pi \rangle + \overbrace{\langle t, \dot{n}_\pi \rangle - \dot{d}_\pi}^{\zeta_\pi} \quad (58)$$

$$= \underbrace{\mathrm{vec}_4(n_\pi)^T J_t}_{J_{t,\pi}} \dot{q} + \zeta_\pi. \quad (59)$$

## V. IMPLEMENTATION ASPECTS

Beyond the mathematical derivations, this section focuses on implementation aspects regarding two topics. First, we show a feasible algorithm for the proposed algorithm, and second, we discuss how to combine the proposed primitives in a relevant manner to prevent collisions between tools and complex anatomy in robot-aided surgery.

### A. Computational algorithm

Without loss of generality, the algorithm to implement the proposed dynamic active constraints for a single robotic system using vector-field inequalities method is shown in Algorithm 1.

### B. Combination of primitives

The proposed framework relies on the combination of primitives to describe the interactions with the environment with a sufficient complexity for a given task. In minimally invasive surgery and microsurgery, possibly the most important requirement is the prevention of collisions between tool shafts and anatomy. Hence, we will briefly describe one way to model that interaction using the proposed geometrical primitives in this section.

For the purposes of endonasal surgery, describing the endonasal safe region for the tool shaft is paramount. For this purpose, the required complexity should be defined beforehand. As far as the safety of the patient is concerned, the safe region has to be conservative. However, it cannot be too

**Algorithm 1** Dynamic active constraints with vector field inequalities control for a single robot.

1: $\tau \leftarrow$ sampling time
2: $\eta_d \leftarrow$ vector field inequalities gain
3: $\eta \leftarrow$ control gain
4: **while** not stopped **do**
5:     $t \leftarrow$ current time
6:     $\underline{q}(t) \leftarrow$ robot's joint position
7:     $\underline{x}_d(t) \leftarrow$ desired pose
8:     $\underline{x}(q) \leftarrow$ robot's pose from FKM
9:     $\underline{\tilde{x}} = \underline{x} - \underline{x}_d$
10:    $J(q) \leftarrow$ task Jacobian
11:    $O = \| J\dot{q} + \eta \, \text{vec}_8 \, \underline{\tilde{x}} \|^2$
12:    **for each** robot entity $\underline{a}$ **do**
13:       $\underline{a}(q) \leftarrow$ robot entity
14:       $J_a(q) \leftarrow$ robot entity's Jacobian
15:    **end for**
16:    **for each** environmental entity $\underline{b}$ **do**
17:       $\underline{b}(t) \leftarrow$ environmental entity
18:       $\underline{\dot{b}}(t) \leftarrow$ environmental entity's velocity
19:    **end for**
20:    $W \leftarrow$ empty matrix
21:    $w \leftarrow$ empty vector
22:    **for each** robot–environmental entity pair **do**
23:       $d_{\text{safe},a,b}(t) \leftarrow$ a-b pair's safe distance
24:       $\dot{d}_{\text{safe},a,b} \leftarrow$ a-b pair's safe distance derivative
25:       $d_{a,b}(\underline{a},\underline{b}) \leftarrow$ a–b pair's distance
26:       $J_{a,b}(\underline{a}, J_a, \underline{b}) \leftarrow$ a–b pair's Jacobian
27:       $\zeta_b(\underline{a},\underline{b},\underline{\dot{b}}) \leftarrow$ environmental entity's residual
28:       **if** Stay outside restricted region (Constraint 18) **then**
29:          $\tilde{d}_{a,b} \leftarrow d_{\text{safe},a,b} - d_{a,b}$
30:          $\zeta_{\text{safe},a,b} \leftarrow \dot{d}_{\text{safe},a,b} - \zeta_b$
31:          AppendRow($W$, $+J_{a,b}$)
32:          Append($w$, $\eta_d \tilde{d}_{a,b} + \zeta_{\text{safe},a,b}$)
33:       **else**  ▷ Stay outside restricted region (Constraint (19))
34:          $\tilde{d}_{a,b} \leftarrow d_{a,b} - d_{\text{safe},a,b}$
35:          $\zeta_{\text{safe},a,b} \leftarrow \zeta_b - \dot{d}_{\text{safe},a,b}$
36:          AppendRow($W$, $-J_{a,b}$)
37:          Append($w$, $\eta_d \tilde{d}_{a,b} - \zeta_{\text{safe},a,b}$)
38:       **end if**
39:    **end for**
40:    $\dot{q} \leftarrow$ SolveQuadraticProgram($O, W, w$)
41:    SendRobotVelocity($\dot{q}$)
42:    SleepUntil($t + \tau$)
43: **end while**

conservative; otherwise, it might be impossible to perform the required task. For example, as shown in Fig. 5, by adding there robot line-to-point constraints, a complex anatomical constraint can be conservatively protected. The designer might increase or decrease the number of line-to-point constraints according to the needs of a given task.

## VI. SIMULATION AND EXPERIMENTS

This section presents a simulation and experiments using two robotic manipulators (VS050, DENSO Corporation, Japan), each with six DOFs and equipped with a rigid 3.0-mm-diameter tool. The physical robotic setup shown in Fig. 9 is the realization of our initial concept to deal with minimally invasive transnasal microsurgery [2].

We first present simulation results to evaluate the behavior of the proposed technique under different conditions and parameters. In addition, to elucidate real-world feasibility, we present three experiments with the physical robotic system.

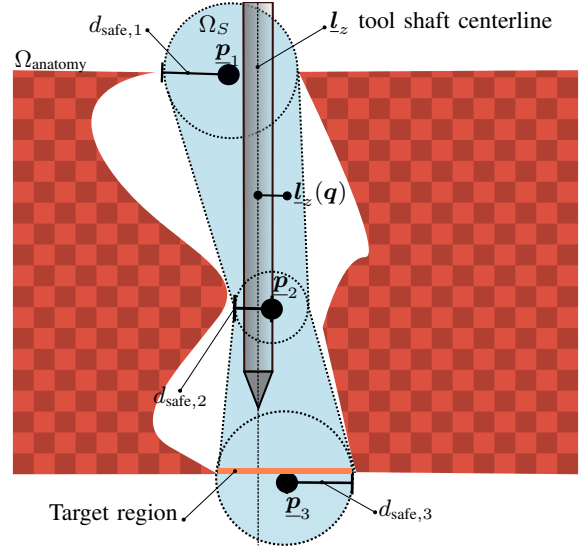

Fig. 5. Example of the combination of point-to-line constraints to generate a 3D safe zone, $\Omega_{\text{safe}}$, which is conservative with respect to the anatomy, $\Omega_{\text{anatomy}}$. Given three points $\underline{p}_1, \underline{p}_2, \underline{p}_3$, their distances are constrained with respect to the tool shaft's centerline, $\underline{l}_z(q)$. If the number of points is increased, the anatomy can be described less conservatively.

The first one is used to evaluate the effect of the parameter $\eta_d$ in the vector-field inequality in (16); the second one is used to evaluate the use of several simultaneous active constraints to avoid collisions with an anatomically correct head model while a pair of robots automatically track a prescribed trajectory; and the last one is used to evaluate the system behavior under teleoperation and tool–tissue interactions with a flexible tissue phantom.

Both simulations and experiments used the same software implementation on a single computer running Ubuntu 16.04 x64 with a kernel patch to Linux called CONFIG_PREEMPT_RT[9] from Ingo Molnar and Thomas Gleixner. In our architecture, the optimization algorithm was solved using the real-time scheduling SCHED_DEADLINE[10] in the user space. In our in-house testing, SCHED_DEADLINE was far superior to SCHED_RR or SCHED_FIFO and provided hard-real-time performance. The b-Cap[11] protocol was used to communication with robots over UDP with sampling time of 8 ms.[12] ROS Kinetic Kame[13] was used for the interprocess communication of non-real-time events, namely communication with VREP and data logging. Furthermore, the dual-quaternion algebra and robot kinematics were implemented using DQ Robotics[14] in C++, constrained convex optimization was implemented using IBM ILOG CPLEX Optimization Studio[15] with Concert Technology, and VREP [44] was used for visualization only.

---

[9]https://rt.wiki.kernel.org/index.php/Frequently_Asked_Questions
[10]http://man7.org/linux/man-pages/man7/sched.7.html
[11]densorobotics.com/products/b-cap
[12]Each DENSO VS050 robot controller server runs at 125 Hz as per factory default settings.
[13]http://wiki.ros.org/kinetic/Installation/Ubuntu
[14]https://dqrobotics.github.io/
[15]https://www.ibm.com/bs-en/marketplace/ibm-ilog-cplex

In the experiments, the tool-tip position with respect to the robot end effector was calibrated by using a visual tracking system (Polaris Spectra, NDI, Canada) through a pivoting process.[16] The visual tracking system was also used to calibrate one robot base with respect to the other.

### A. Simulation: Shaft–shaft active constraint

In the first simulation, two robots (R1 and R2) were positioned on opposite sides of the workspace. The robots were initially positioned so that their tool shafts cross when viewed from the $xz$-plane. Both robots were commanded to simultaneously move their tools along the $y$-axis. The tools first moved along the same trajectory along the negative $y$-axis direction such that R1 followed R2 with a slight offset during $t \in [0, 2 \text{ s})$; then, both robots maintained simultaneous motion but changed their direction along the $y$-axis so that R2 followed R1 during $t \in [2 \text{ s}, 4 \text{ s})$, moved toward each other during $t \in [4 \text{ s}, 6 \text{ s})$, and finally moved away from each other during $t \in [6 \text{ s}, 8 \text{ s}]$. The reference trajectory required the tool orientations to remain constant. The tools were simulated as 3.0-mm-diameter cylindrical shafts.

In order to study the dynamic active constraint to prevent collision a between shafts, we consider three different ways of generating the control inputs. In the first one, the control input is generated by the solution of Problem 8; therefore, as the robot does not take into account any obstacle and is unaware of its surroundings, we say it is *oblivious*. In the second and third ones, the control input is generated by the solution of Problem 10 with Constraint 18 and (48), (49), and (50). However, in the second one, the residual is not used; thus, it is a *static-aware* robot. That is, the robot considers its surroundings but does not take into account obstacle kinematics. In the third one, a robot controlled with the full proposed dynamic active constraints including the residual is called a *kinematics-aware* robot. If both robots are kinematics-aware, the centralized Problem 12 was used to solve for both robots simultaneously. With these definitions, we examine every unique combination of states between robots for a convergence rate $\eta = 50$, an approach gain $\eta_d \in \{0, 1, 2, 3, 4, 5, 6, 7, 8.5, 10\}$, and a sampling time $\tau = 8$ ms. A *static-aware* robot is what could be achieved without using the proposed residuals; therefore, it is equivalent to [32]. Hence, the results of these simulations are appropriate to compare our initial results presented in [32] to the ones reported in this paper.

The design parameter $\eta_d$ corresponds to the fastest distance decay allowed between two moving robotic systems; therefore, in a system where both robots are kinematics-aware, have the same approach gain $\eta_d$, and move directly towards each other, the result of the optimization will assign a gain of $\eta_d/2$ for each robot in the most restrictive case. Thus, to provide a fair comparison in the same scenario, when both robots are static-aware, the approach gain $\eta_d$ was divided in half.

*1) Results and discussion:* The performance for $\eta_d = 2$ in terms of the trajectory tracking error and minimum separation

[16]The pivoting process consists in adding a marker to the tool, then pivoting the tool tip about a divot and recording the marker coordinates to obtain the relative translation between marker and tool tip. This process is available in the manufacturer's software.

TABLE III
PERFORMANCE METRICS FOR SIMULATION A WITH $\eta_d = 2$.

| R1 | R2 | $\int \|\tilde{\boldsymbol{x}}_1\|_2$ | $\int \|\tilde{\boldsymbol{x}}_2\|_2$ | $\int \|\tilde{\boldsymbol{x}}_1\|_2 + \int \|\tilde{\boldsymbol{x}}_2\|_2$ | collision |
|---|---|---|---|---|---|
| o | o | 0.1484 | 0.1488 | 0.2973 | yes |
| o | s | 0.14845 | 8.1995 | 8.3479 | yes |
| s | o | 7.93310 | 0.1488 | 8.0819 | yes |
| o | k | 0.1484† | 6.1087* | 6.2571 | **no** |
| s | s | 3.22129 | 5.7271 | 8.9483* | **no** |
| s | k | 4.2704 | 3.2258 | 7.4962 | **no** |
| k | o | 6.1498* | 0.1488† | 6.2986 | **no** |
| k | s | 1.97851 | 5.9864 | 7.9649 | **no** |
| k | k | 3.0947 | 3.1159 | 6.2107† | **no** |

Robots R1 and R2 can be oblivious (*o*), static aware (*s*), and kinematics aware (*k*). Moreover, $\tilde{\boldsymbol{x}}_1$ is the trajectory error of R1, and $\tilde{\boldsymbol{x}}_2$ of R2. Lastly, we considered a collision whenever the distance between shafts was smaller than 3 mm. Considering only cases with no collision, values with † are the best values in each column and values with * are the worst values in each column.

distance is summarized in Table III. The continuous behavior can be seen in Fig. 6. Since the simulations for other approach gains show a similar trend, they are omitted.

When both robots were oblivious, the tool shafts collided as no constraints were enabled. When one robot was static-aware and the other robot was oblivious, a snapshot vision of the world was not sufficient to avoid collisions. In the remaining cases, collisions were effectively avoided.

When both R1 and R2 were static-aware robots, the sum of the tracking errors for R1 and R2 was the largest. As the robots considered each other as an instantaneously static obstacle, they slowed down to avoid collisions, even when the other robot was effectively moving away. This behavior is shown in Fig. 6 for R1 between 0 and 2 s, and for R2 between 2 and 4 s. These results indicate that tracking error is suboptimal when using only static awareness, even though such a configuration might be reliable to prevent collisions if both arms implement it. For instance, if one of the tools is directly manipulated by a surgeon, it may be difficult to estimate its velocity, but as shown by our results, only the tool's position is needed to prevent collisions since the surgeon will use their static knowledge to avoid dangerous interactions between tools.

When one robot was kinematics-aware and the other was oblivious, the oblivious robot followed its trajectory with minimum error, as it was unaware of the existence of the other robot. The kinematics-aware robot used knowledge of the kinematics to fully undertake evasive maneuvers. Such results indicate that this configuration may be useful in practical scenarios in which one robot has a clear priority over the other. For instance, this configuration would be useful when a human collaborator handles a sensorized tool for which the velocity and position can be obtained.

When both robots were kinematics-aware, they had the best combined trajectory tracking for cases in which no collision happened. Indeed, this happened because such a configuration provides a more suitable problem description, which is to minimize the sum of the trajectory errors under the anti-collision constraints. In fact, adding the residual information in Problem 12 allows the system to optimize the approach velocity when one entity is moving away. Such behavior justifies the implementation of dynamic active constraints whenever feasible.

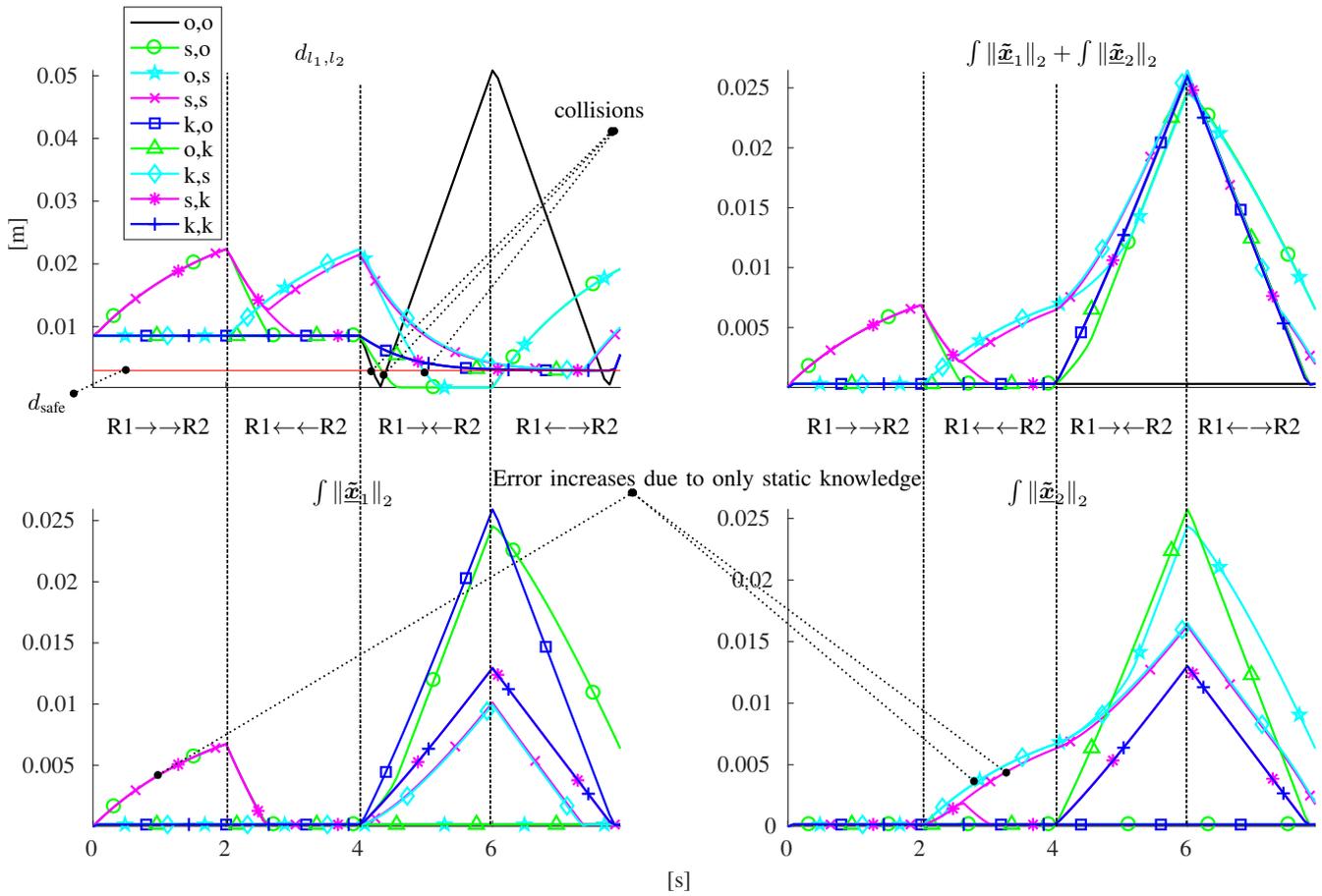

Fig. 6. Simulation A. The vertical dashed bars represent the transition to a different phase, which are named as: R1 →→ R2 when R1 moved in the direction of R2, which moved away; R1 ←← R2 when R2 moved in the direction of R1, which moved away; R1 →← R2 when the robots moved towards each other; and R1 ←→ R2 when the robots moved away from each other. Collisions only happened in the set $\{\{o,o\}, \{o,s\}, \{s,o\}\}$.

### B. Experiment A: Plane active constraint

TABLE IV
CONTROL PARAMETERS FOR THE EXPERIMENTS.

| Experiment | $\eta$ | $\eta_d$ | $\lambda$ |
|---|---|---|---|
| A | 50 | $\{0, 0.25, 1, 4, 16\}$ | 0 |
| B | 300 | 2 | 0.1 |
| C | 50 | $\{1, 2, 4, 8\}$ | 0.1 |

Experiments A and C had trials with different values of $\eta_d$, one with each value in the set.

In this experiment, we evaluated the behavior of the robot when its tool tip was commanded to reach a location beyond a static plane boundary while studying different approach gains for the point-to-plane active constraint. The experimental setup is shown in Fig. 7. The instrument consisted of a 3-mm-diameter hollow aluminum shaft.

The initial robot configuration was $\boldsymbol{q}(0) = \begin{bmatrix} 0 & \pi/6 & \pi/2 & 0 & 0 & -\pi/6 & 0 \end{bmatrix}^T$ rad, and the robot was commanded to move down 45 mm along the world $z$-axis, while the rotation was fixed with the tool tip pointing downward. The constraint consisted of a static plane orthogonal to the $z$-axis and positioned 25 mm below the initial tool tip position. Hence, a collision was expected after the tool tip moved 25 mm downwards. The initial condition and reference trajectory were the same for all executions.

In the first execution, there was no plane constraint. To study the effects of different active constraint gains, we chose gains in the set $\eta_d = \{0, 0.25, 1, 4, 16\}$ in each execution[17].

*1) Results and discussion:* As shown in Fig. 8, when the active constraint was not activated, the end effector moved through the plane, and the trajectory had the minimum tracking error, as nothing constrained the motion.

When $\eta_d = 0$, the end effector was not allowed to approach the plane. As the vector-field-inequality gain increased, the end effector was allowed to approach faster, as shown by the exponential behavior in Fig. 8. Even with an increase in the tracking error norm, as shown in the bottom graph in Fig. 8, the end effector did not cross the plane for any value of the approach gain, which is predicted by theory, and there was no sudden activation of the joints, as would happen if position-based constrained optimization [12], [13] was used.

For this experiment (i.e., six executions, each one corresponding to one approach gain), the full control loop had an average computation time of $3.9\,\mathrm{ms}$ with a standard deviation of $0.6804\,\mathrm{ms}$ for nonoptimized C++ code.

### C. Experiment B: Constrained workspace manipulation

This section presents experiments in a mockup of the constrained neurosurgical setup shown in Fig. 9 for dura mater

---
[17]Refer to the video attached to this submission.

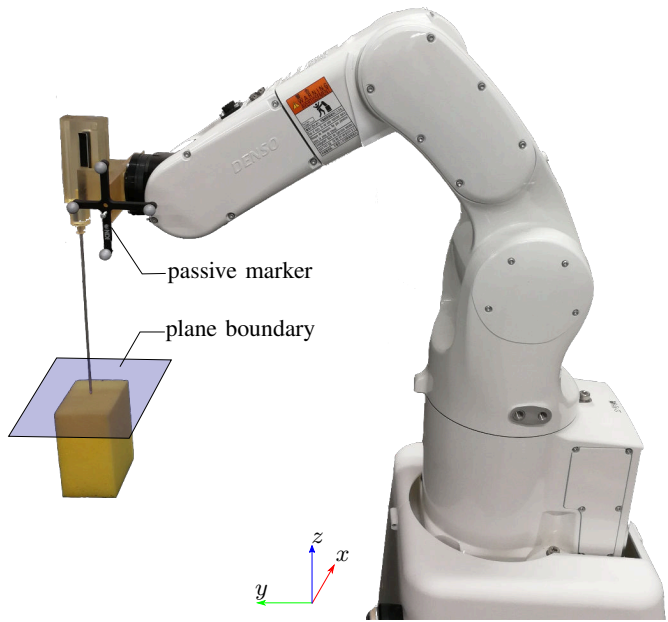

Fig. 7. Setup for experiment A, described in Section VI-B. A single robot was used and the passive marker was used to calibrate the tool tip. The plane was chosen so that the tool tip did not touch the yellow sponge cube.

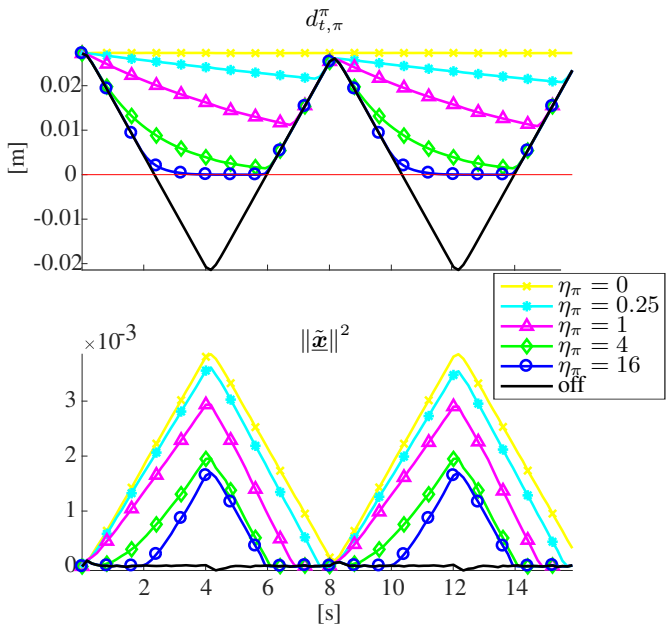

Fig. 8. Results of six executions in Experiment A, where a static plane generated a forbidden region, as described in Section VI-B. The distance between the end effector and the plane is shown on *top* and the 2-norm of the pose error, as the robot tracked the desired trajectory, is shown on the *bottom* graph. The distance limit is represented by the red line, and whenever the active constraint was enabled the end effector did not go through the plane.

manipulation using a transnasal approach.[18]

[18]In pituitary gland resection, long thin tools are inserted through the nasal cavity to reach the sphenoid sinus. To increase workspace in the nasal cavity, some turbinates (i.e., protruding bones in the nasal cavity with several physiological uses, which can be removed to some extent without major impact on the patient's health) are removed by an ear, nose, and throat (ENT) specialist. After the clean-up done by the ENT specialist, a neurosurgeon takes place and removes a thin bone shell protecting the sphenoid sinus, exposing the dura mater. After being exposed, the dura mater is cut, the pituitary gland is resected, and the dura mater is then sutured.

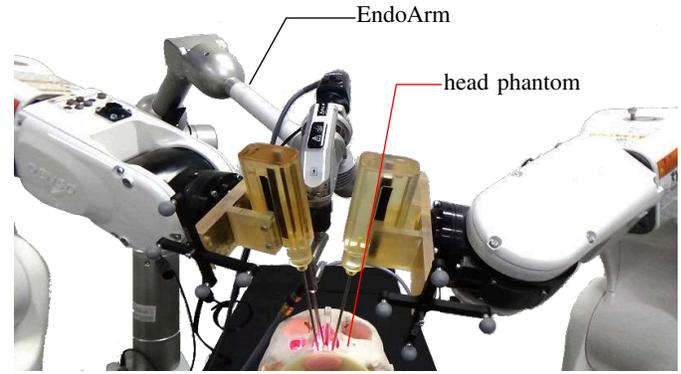

Fig. 9. Setup for the experiment described in section VI-C.

The mockup consisted of a pair of manipulators positioned on each side of a 3D-printed head model (M01-SET-TSPS-E1, SurgTrainer, Ltd, Japan). The head model was customized to our needs by adding the target dura mater. For high-definition vision, we used a $30°$ endoscope (Olympus, Japan) attached to a manually operated endoscope holder (EndoArm, Olympus, Japan). The dura mater and surgical tool tips along with the desired trajectories are shown in Fig. 10. As we used a mock tool that consisted of a 3-mm-diameter hollow aluminum shaft without any actuation, the whole system had six DOFs for each robotic manipulator. Furthermore, only the tool tip trajectory was controlled, hence the robot was redundant to the task, having a surplus of three DOFs for each robot.

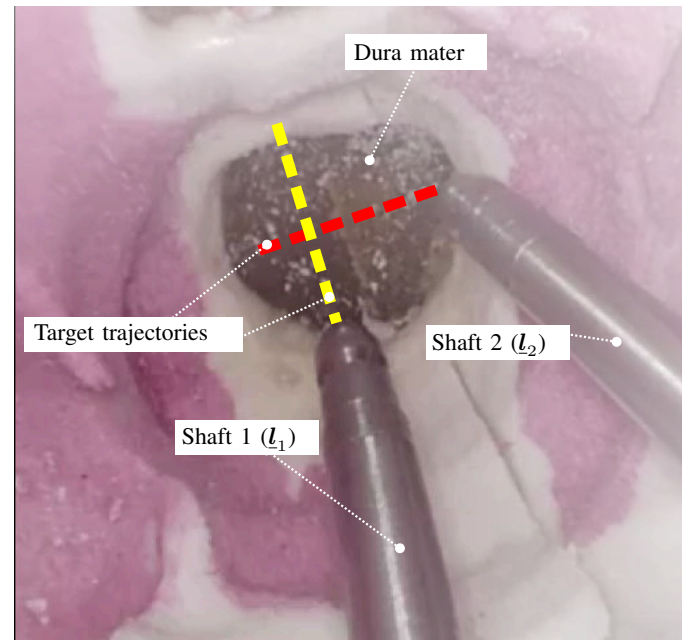

Fig. 10. Setup for the experiment described in Section VI-C.

Although this experiment was devised to be as close as possible to a clinical scenario, it does not aim to be clinically relevant by itself as it would require several considerations outside the scope of this paper. Therefore, the purpose of this experiment is to evaluate if the proposed active constraints can be used in a realistic environment and to answer the following research questions:

1) Do the added disturbances of physical robotic systems destabilize the control law?
2) Can a complex environment be effectively modeled by

using the proposed primitives?
3) Is there a feasible method to calibrate the robotic systems to obtain a reasonable accuracy?

These questions define whether collisions can be effectively prevented by using the kinematic information of both the robots and workspace.

The target procedure, as surgical procedures in general, has intricate subtasks [45]. A treatise on the entire procedure might require the online calculation of the procedure workflow, which is an active area of research [46], but would be outside of the scope of this work. In this work, we focus on evaluating our technique on the most challenging surgical subtask for endonasal surgery, in which, according to our medical staff, robotic aid would have the highest impact. More specifically, after the dura matter (membrane) is exposed, the surgeon has to manipulate it through complex precise motions. In this subtask, carrying out the procedure requires:

1) *Collision-free robot motion.* Preventing collisions with the anatomy is paramount. In fact, if there are no active constraints, using a robot to aid in such procedure is impossible due to collisions with the anatomy;
2) *Safe insertion–retraction of the robotic tools from the patient.* In this type of surgery, the retraction and insertion of tools is frequent, in order to clean instruments, remove debris, etc.

With these requirements in mind, three groups of active constraints were used. The first group was composed of the line-to-point constraints of the robot (see Fig. 11). A pair of points were placed in the entrances of the nostrils with positions $p_{nl}$ and $p_{nr}$, another pair at the ends of the sinuses $p_{el}$ and $p_{er}$, and a last point positioned at the center of the dura matter, $p_m$. Moreover, the left robot's tool shaft was collinear with the Plücker line $\underline{l}_1 \triangleq \underline{l}_1(q_1)$, and the right robot's tool shaft was collinear with $\underline{l}_2 \triangleq \underline{l}_2(q_2)$. The distances between shafts and points were constrained to specify safe zones for tools in order to prevent collisions between the tool shafts and the head model. In this way, the first group of constraints related to the first robot were chosen as

$$\mathcal{C}^1_{l_1, p_{nl}} \triangleq J_{l_1, p_{nl}} \dot{q}_1 \leq \eta_{l_1, p_{nl}} \tilde{D}_{l_1, p_{nl}}, \quad (60)$$

$$\mathcal{C}^2_{l_1, p_{el}} \triangleq J_{l_1, p_{el}} \dot{q}_1 \leq \eta_{l_1, p_{el}} \tilde{D}_{l_1, p_{el}},$$

$$\mathcal{C}^3_{l_1, p_m} \triangleq J_{l_1, p_m} \dot{q}_1 \leq \eta_{l_1, p_m} \tilde{D}_{l_1, p_m}, \quad (61)$$

and the ones related to the second robot were chosen as

$$\mathcal{C}^4_{l_2, p_{nr}} \triangleq J_{l_2, p_{nr}} \dot{q}_2 \leq \eta_{l_2, p_{nr}} \tilde{D}_{l_2, p_{nr}}, \quad (62)$$

$$\mathcal{C}^5_{l_2, p_{er}} \triangleq J_{l_2, p_{er}} \dot{q}_2 \leq \eta_{l_2, p_{er}} \tilde{D}_{l_2, p_{er}},$$

$$\mathcal{C}^6_{l_2, p_m} \triangleq J_{l_2, p_m} \dot{q}_2 \leq \eta_{l_2, p_m} \tilde{D}_{l_2, p_m}, \quad (63)$$

where both squared distance functions and Jacobians were chosen as in (33) and (34), respectively. Since the points attached to the head model did not change during the experiments, the corresponding residuals are zero.

The second group of constraints was composed of four plane-to-point constraints for the robot, as shown in Fig. 12. The plane attached to the left robot is given by $\underline{\pi}_1 \triangleq \underline{\pi}_1(q_1)$ and the points attached to the right robot are given by $p_{R2,i} \triangleq p_{R2,i}(q_2)$, $i \in \{1, 2, 3, 4\}$. In order to constrain the distance between these points and the plane (thus preventing collision between the tool modules outside the patient), the four constraints of the second group were chosen as

$$\mathcal{C}^{i+6}_{R_2, i} \triangleq -J_{\pi_1, R2, i} \dot{q}_1 - J_{R2, i, \pi_1} \dot{q}_2 \leq \eta_{\pi_1, R2, i} \tilde{D}_{\pi_1, R2, i}.$$

Lastly, a combination of line-to-point and robot line-to-line constraints for the robot were used to prevent collisions between the shafts, each modeled as a line with the tool tip as one endpoint. Let the tool tip positions be given by $t_1 \triangleq t_1(q_1)$ and $t_2 \triangleq t_2(q_2)$. Each semi-infinite cylinder was therefore $c_1 \triangleq c_1(t_1, \underline{l}_1, d_{\text{safe1}})$ and $c_2 \triangleq c_2(t_2, \underline{l}_2, d_{\text{safe2}})$, in which $d_{\text{safe1}}$ and $d_{\text{safe2}}$ represent their radii. This was implemented as two constraints:

$$\mathcal{C}^{11}_{c_1, c_2} \triangleq \begin{cases} -J_{t_1, l_2} \dot{q}_1 - J_{l_2, t_1} \dot{q}_2 \leq \eta_{t_1, l_2} \tilde{D}_{t_1, l_2} & \text{if } \text{proj}_{\underline{l}_2}(t_1) \in c_2, \\ -J_{t_2, l_1} \dot{q}_2 - J_{l_1, t_2} \dot{q}_1 \leq \eta_{t_2, l_1} \tilde{D}_{t_2, l_1} & \text{if } \text{proj}_{\underline{l}_1}(t_2) \in c_1, \\ \mathbf{0}^T \dot{q}_1 + \mathbf{0}^T \dot{q}_2 \leq 0 & \text{otherwise}, \end{cases} \quad (64)$$

in which $\text{proj}_{\underline{l}}(p)$ is the projection of $p$ onto $\underline{l}$ and

$$\mathcal{C}^{12}_{c_1, c_2} \triangleq \begin{cases} -J_{l_1, l_2} \dot{q}_1 - J_{l_2, l_1} \dot{q}_2 \leq \eta_{l_1, l_2} \tilde{D}_{l_1, l_2} & \text{if } \text{proj}_{\underline{l}_1}(\underline{l}_2) \in c_2 \\ & \text{and } \text{proj}_{\underline{l}_2}(\underline{l}_1) \in c_1, \\ \mathbf{0}^T \dot{q}_1 + \mathbf{0}^T \dot{q}_2 \leq 0 & \text{otherwise}. \end{cases} \quad (65)$$

Arbitrary collisions between cylinders is the topic of extensive research [47], [48], and we focused only on relevant collisions for this experiment.

In total, twelve constraints were used in the control algorithm, whose control input was generated by the solution of Problem 12. Similar to real surgery, the endoscope was positioned in such a way to not disturb the motion of the tool, therefore no specific active constraints were required for the endoscope.

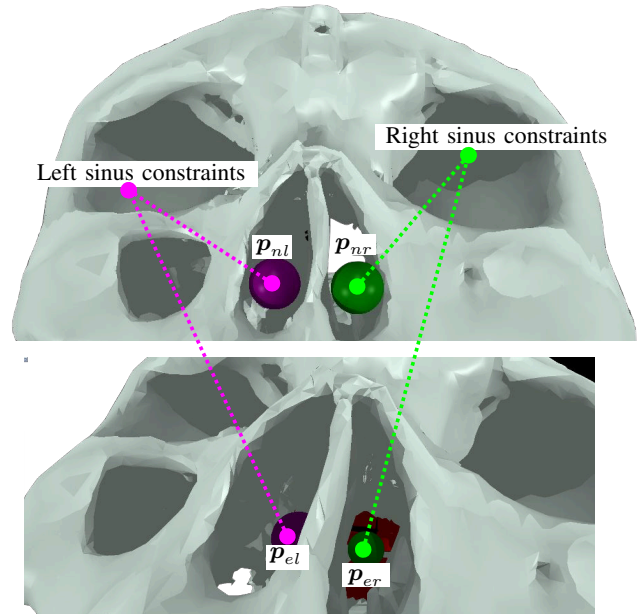

Fig. 11. 3D computer model of the compliant pivoting points used to prevent collisions between the tool shafts and the anatomical model. The points were fixed with respect to the head model, and the distance between points and tool shafts was actively constrained.

The experiment was divided into three parts, all with the same initial configuration, as shown in Fig. 10. The trajectories consisted of a straight line connecting the start and end points, which were automatically tracked by the tool tips. The first part consisted of letting the right tool be static and moving the left tool from the initial bottom position to the top along the yellow trajectory and returning. The second part consisted of letting the left tool be static while moving the right tool from the initial right position along the red trajectory to its leftmost point, and returning. The last part consisted of moving both tools simultaneously along their predefined trajectories. The same set of constraints were used in all three parts. The two first parts were used as ground-truth, as collisions were not expected between tools. The last part required the tools to follow a trajectory that would cause a collision unless the manipulators performed evasive maneuvers.

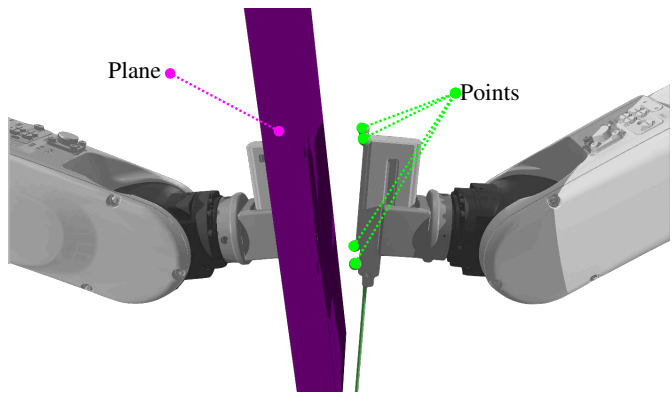

Fig. 12. 3D computer model of the constraints used to prevent collisions between the forceps modules. A dynamic plane was rigidly attached to the left robot whereas four points were rigidly attached to the right robot. The distance between the plane and the points was actively constrained.

The control parameters are shown in Table IV.

*1) Results and discussion:* A representative trajectory for each of the three runs is shown in Fig. 13. As the manipulator robots have rigid links and high-accuracy calibration, all executions effectively yielded the same results. A sequence of snapshots is shown in Fig. 13, showing how collision avoidance was performed.[19]

Fig. 14 shows the signed distances between the actively constrained elements, which were calculated using the robots' kinematic models. A negative signed distance means that a collision happened. Possible collisions with the head model were visually inspected.

As shown in Fig. 13, when only one tool moved, the desired trajectory was tracked without major deviations. Using this effective trajectory as a point of comparison, Fig. 13 also shows the required deviation when both tools moved simultaneously so that no collision would happen.

The distances measured using the robots' kinematics, as shown in Fig. 14, show that the disturbances and noise caused by the experimental system did not affect the usability of the proposed framework. This experimentally validates the results initially studied in Section VI-A.

Decomposing the workspace into primitives provided safety without hindering tool motion. In contrast with techniques that require full knowledge of the anatomy [13], [28], the proposed methodology can be used intraoperatively and is computationally efficient. For instance, the surgeon can define regions inside the workspace in succession to define where the compliant pivoting points should be located and build the safe regions for the tools.

[19]Refer to the video attached to this submission.

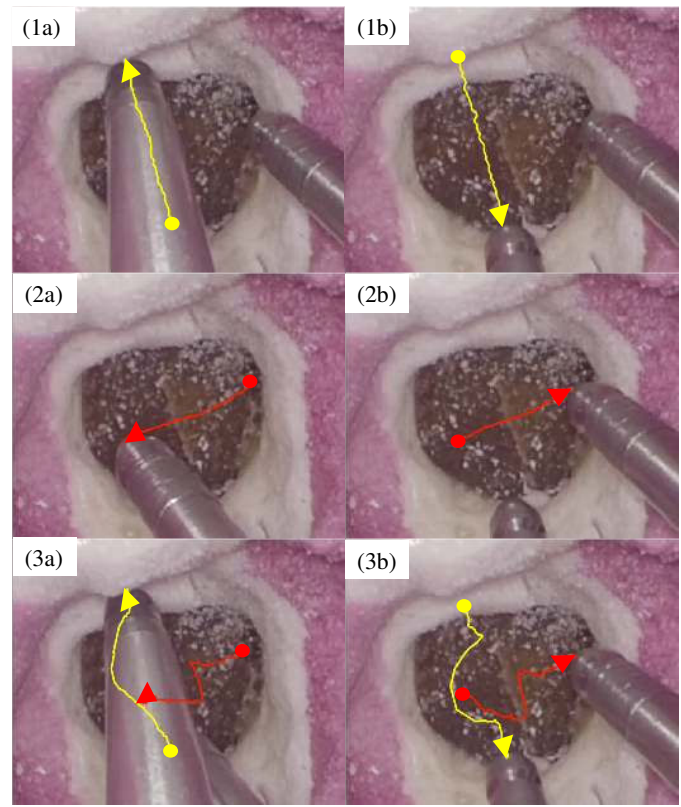

Fig. 13. Tracking results for the three experimental runs described in Section VI-C. Each trajectory was divided into two parts for clarity, and the executed trajectory in each case was obtained by manually marking tool tip pixels every ten video frames. In each trajectory, a circle marks its beginning and an arrowhead marks its end. Figures *1a* and *1b* show the trajectories executed by the left tool while the right tool was fixed. Conversely, Figs. *2a* and *2b* show the trajectories executed by the right tool while the left tool was fixed. Lastly, Figs. *3a* and *3b* show the trajectories executed when both tools moved simultaneously, illustrating how, in order to avoid collisions, both tools deviated from the prescribed trajectory.

A major point of concern is the level of accuracy of the calibration since it directly affects the kinematic control laws such as the one proposed in this work. In these experiments, which were sufficiently complex, a relatively common visual tracking system was sufficient to provide a reasonable level of accuracy, but some inaccuracy was still visible. The inaccuracy caused by the system can be qualitatively seen when comparing the expected simulated motion and actual robot motion. The major source of discrepancies in this case was the calibration of the tool shaft axes, which was not directly done. This is a topic of ongoing research in our group.

In summary and in reply to the research questions imposed at the beginning of this section, disturbances inherent to physical robotic systems did not destabilize the control law, the complex endonasal environment was effectively modeled using the proposed primitives, and a relatively common visual tracking system was sufficient to provide a reasonable level of accuracy.

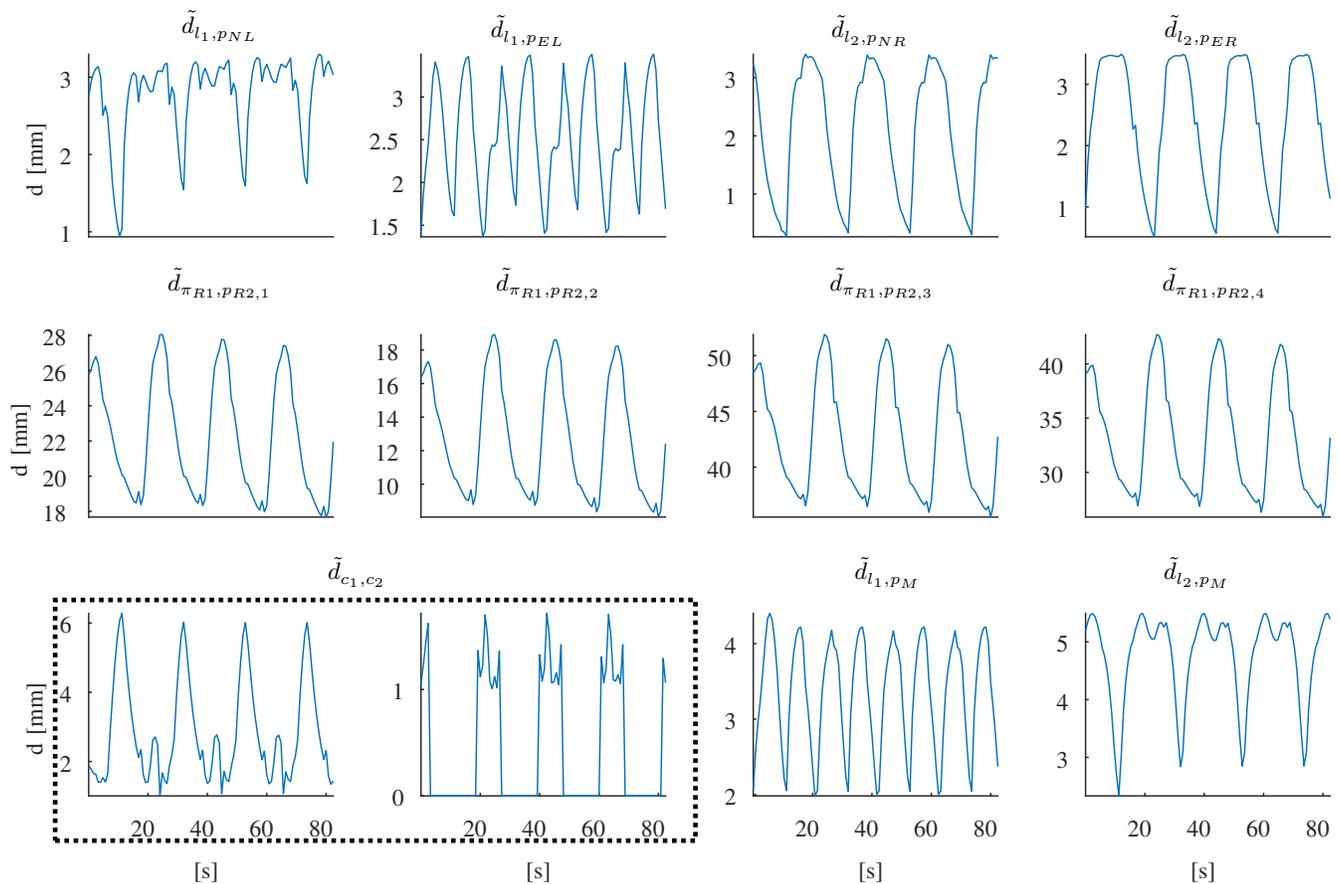

Fig. 14. Signed distances between primitives in the third part of Experiment B, in which both robots moved. Four sequential repetitions of the same motion are shown. In each plot, positive signed distance means that the robot is in the safe zone; as a negative value would mean that the constraint was violated, all curves show that the constraints were respected throughout the whole experiment.

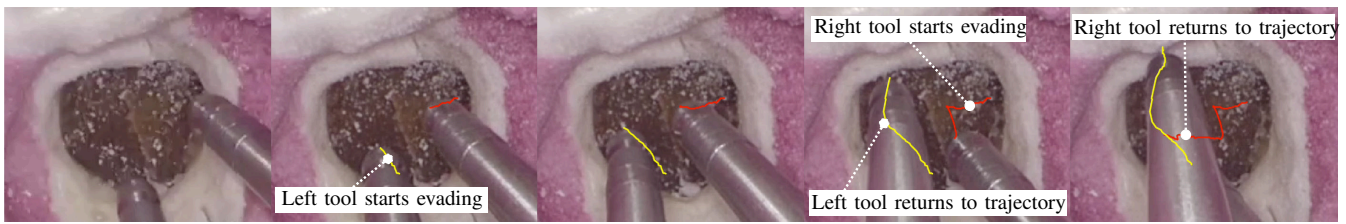

Fig. 15. Snapshots of the tool evasion in the third automatic tracing experiment, in which both robots moved. The left tool uses the dynamic knowledge of the right tool to initiate the evasive maneuvers before getting close to it.

### D. Experiment C: Deformable tissue manipulation during teleoperation

This section presents the results of an experiment to evaluate the performance of the proposed technique during tool–tissue interactions during teleoperation. The robotic system used for this experiment was the same as that presented in Section VI-C.

The pair of robots were positioned on each side of a realistic head model[20] based on human anatomy with added soft tissues, and the objective for this procedure was cutting a membrane in a more challenging location than the one in Section VI-C, which was observed through a 70° endoscope (Olympus, Japan) attached to a manually operated endoscope holder (EndoArm, Olympus, Japan). The membrane model of this experiment was the same material used in latex surgical gloves, which are commonly employed by practicing surgeons. The left arm had the same hollow aluminum shaft used in the experiment in Section VI-C. The right tool had a microsurgical handle (NF-200BA, FEATHER, Japan) attached to it and, at the handle's end, we attached a microsurgical knife (K-6010, FEATHER, Japan). The arm tips and relative pose were calibrated in a similar fashion to the experiment in Section VI-C using passive markers and an upgraded visual tracking system (Polaris Vega, NDI, Canada).

In this experiment, the right tool's desired position was sent through a master interface (Phantom Premium 1.0, 3D

---

[20]This head model is being developed in a cooperation between the University of Tokyo, the University of Nagoya, the University of Tokyo Hospital, Meijo University, and the National Institute of Advanced Industrial Science and Technology, in the context of the ImPACT Bionic Humanoids Propelling New Industrial Revolution project funded by the cabined office of Japan.

Systems, USA) at 100 Hz. The task to be performed by the nonspecialist operator was to pierce the membrane's cutline using the right tool's sharp tip. The left tool was positioned at the center of the membrane, and its desired position was fixed throughout the procedure to deliberately obstruct the motion of the right tool. The purpose of this experiment was to evaluate the system's behavior when interacting with soft tissues during teleoperation. Five trials were performed, four of which used the proposed dynamic active constraints with an increasing gain $\eta_d = \{1, 2, 4, 8\}$. The last one was done without the residual, and $\eta_d = 1$, similar[21] to what was proposed in our previous work [32]. Relevant control parameters are shown in Tab. IV.

The model used in this experiment, which was different from the model used in Section VI-C, had the soft tissues of the face and those inside the nostril. Moreover, in order to have a more meaningful experiment, we also targeted a larger membrane so that the cut path was longer.

With these differences, the experiment could be performed with only two sets of active constraints. The first set was the two line-to-point constraints of the robot for each arm, which generates a truncated-cone safe region for each tool shaft, namely, Constraints 60 and 61 for the left tool and Constraints 62 and 63 for the right tool. The second set of constraints were Constraints 64 and 65 to prevent collisions between tools.

*1) Results and discussion:* Snapshots of two of the trials in this experiment ($\eta_d = 1$) with and without residual are shown in Fig. 16.[22]

When the residual was used, the operator was able to command the right tool and perform several incisions along the cut path in the membrane without collisions between tools or between tools and anatomy. The major difference between these trials was how much the left tool moved in order to avoid collisions, as was already evidenced in the simulations (Section VI-A) and experiments (Section VI-B).

The major difference occurred when the same experiment was attempted without the residual, which is equivalent to the technique that we developed in our previous work [32]. In this case, as the left tool was unaware of the right tool's motion, it did not move away when the right tool approached; therefore, the operator could not complete the task.

This experiment showed the capabilities of the proposed framework in a constrained workspace during teleoperation. Future work will focus on taking these results into a more clinical setup. When robots are teleoperated, the surgeon's judgment is necessary to decide which collisions should be avoided and how much autonomy the robotic system can have. For instance, with an actuated tool, the operator may wish that translation is prioritized over rotation. Moreover, in a given scenario, they might want tools to halt instead of autonomously avoiding collisions between each other. For instance, in this experiment, both tools shared the burden of avoiding collisions. In some procedures, however, it may be more intuitive for the operator if the system prioritizes one tool over the other. Lastly, in this experiment, and also as shown in Fig. 13, the tool tip trajectory might be deviated from the desired tool path to prevent collisions. Since such mismatch between desired and actual tool-tip trajectory might affect the control performance, we are currently investigating haptic feedback to inform the operator of such misalignment. The idea we are currently working on is Cartesian feedback, which provides a signal for the operator that is, for instance, proportional to the misalignment and in a suitable direction such that the operator is able to mitigate it.

## VII. CONCLUSIONS

This paper extended our previous work [32] to consider dynamic active constraints. The method can be used to keep the robot from entering a restricted zone or to maintain the robot inside a safe area and is based on a solution that takes into account both equality and inequality constraints. The vector-field inequalities limit the velocity of the robot in the direction of the restricted zone's boundary and leave the tangential velocities undisturbed. The method requires a Jacobian that describes the relation between the time derivative of the distance and the robot joints' velocities as well as a residual that encodes the kinematic information for dynamic objects. We derive the Jacobians and residuals for relations in which both entities can be dynamic, namely, point–point, point–line, line–line, and point–plane. In simulated experiments, a pair of six-DOF robot manipulators equipped with endonasal tool mockups was used to evaluate the behavior of the proposed technique under different conditions and parameters. The results have shown that by using the vector-field inequalities for dynamic objects, the combined trajectory error of the two robotic systems is optimal. Experiments using a real robotic system have shown that the method can be applied to physical systems to autonomously prevent collisions between the moving robots themselves and between the robots and the environment.

The proposed method has shown good results with a reasonably simple calibration procedure using passive markers and a visual tracking system. A current line of research is focused on the integration of dynamic active constraints with tracking algorithms that use endoscopic images, which may increase the overall accuracy of the method. Moreover, clinically oriented studies regarding the teleoperation of a robotic system with a novel four-DOF tool are currently in progress.

## ACKNOWLEDGEMENTS

The authors would like to thank Prof. Taisuke Masuda and Prof. Fumihito Arai (Nagoya University), Dr. Taichi Kin (The University of Tokyo Hospital), Dr. Juli Yamashita (National Institute of Advanced Industrial Science and Technology), and Dr. Akiyuki Hasegawa (Meijo University) for their help in providing us with realistic head models.

---

[21]In fact, in our prior work [32], we used linear programming and the distance Jacobians and distance functions were slightly different. The important difference for our purposes was the lack of residual.

[22]See accompanying video.

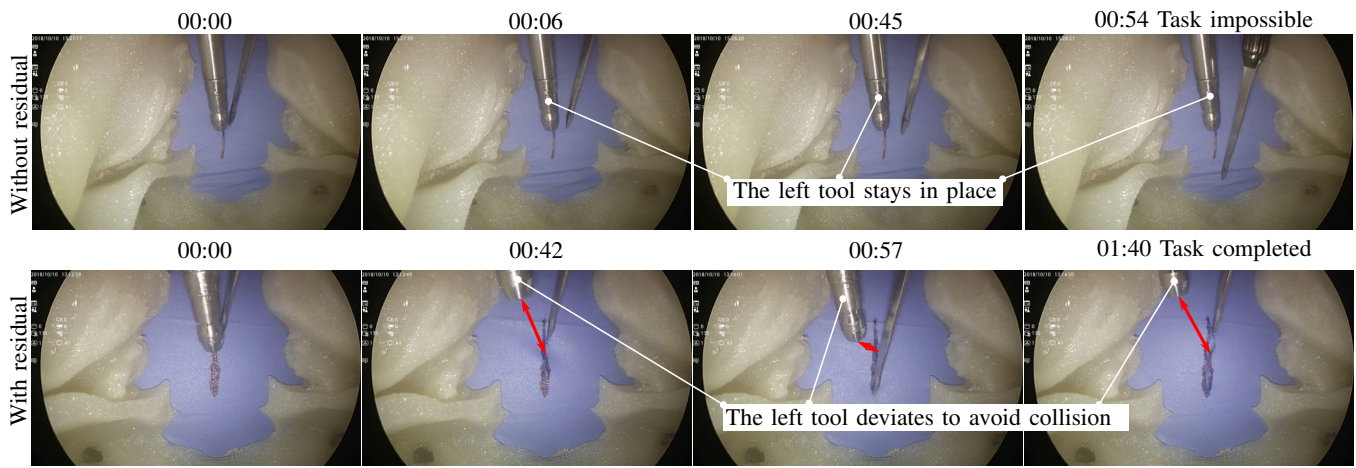

Fig. 16. Results of two of the trials described in Section VI-D. The first row shows the trial with no residual and $\eta_d = 1$ (equivalent to [32]), in which the left tool is unaware of the right tool's motion and therefore it does not move away, therefore it was impossible to perform the cutting task. The second row shows one of the trials in which the residual is used and $\eta_d = 1$. By adding the residual, the centralized optimization problem makes the left tool automatically move away to reduce the overall trajectory tracking error, making it possible to perform the cutting task.

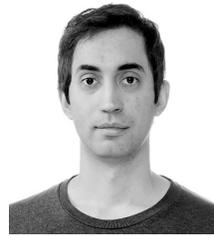

**Murilo Marques Marinho** (GS'13–M'18) received the bachelor's degree in mechatronics engineering and the master's degree in electronic systems and automation engineering from the University of Brasilia, Brasilia, Brazil, in 2012 and 2014, respectively. He received the Ph.D. degree in mechanical engineering from the University of Tokyo, Tokyo, Japan, in 2018. In 2018, he was a Visiting Researcher with the Johns Hopkins University. He is an Assistant Professor with the University of Tokyo from 2019. His research interests include robotics applied to medicine, robot control theory, and image processing.

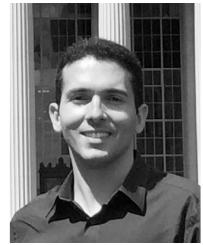

**Bruno Vilhena Adorno** (GS'09–M'12–SM'17) received the BS degree (2005) in Mechatronics Engineering and the MS degree (2008) in Electrical Engineering, both from the University of Brasilia, Brasilia, Brazil, and the PhD degree (2011) from University of Montpellier, Montpellier, France. He is currently a Tenured Assistant Professor with the Department of Electrical Engineering at the Federal University of Minas Gerais, Belo Horizonte, Brazil. His current research interests include both practical and theoretical aspects related to robot kinematics, dynamics, and control with applications to mobile manipulators, humanoids and cooperative manipulation systems.

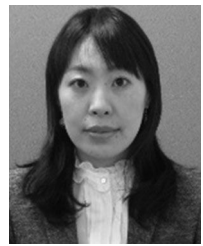

**Kanako Harada** (M'07) received the B.E. and M.E. degrees from The University of Tokyo, Tokyo, Japan, and the Ph.D. degree in engineering from Waseda University, Tokyo, in 1999, 2001, and 2007, respectively. After working with Hitachi, Ltd.,Waseda University, the National Center for Child Health and Development, the Scuola Superiore Sant'Anna, the University of Tokyo, and the Japan Science and Technology Agency (JST), she was appointed as a Associate Professor at The University of Tokyo in 2015. Her research interests include medical and surgical robotics.

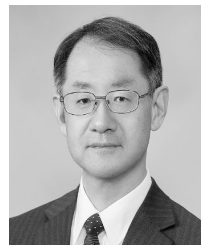

**Mamoru Mitsuishi** (M'93) received the master's and D.E. degrees in mechanical engineering from the University of Tokyo, Tokyo, Japan, in 1983 and 1986, respectively. In 1986, he was a Lecturer with the University of Tokyo, where he was also an Associate Professor in 1989 and a Professor since 1999. From 1987 to 1988, he was a Visiting Researcher with the Fraunhofer Institute for Production Technique and Automation, Stuttgart, Germany. His research interests include computer integrated surgical systems and manufacturing systems.